\documentclass[conference]{IEEEtran}
\usepackage{times}

\usepackage[numbers]{natbib}
\usepackage{multicol}
\usepackage[bookmarks=true]{hyperref}
\usepackage{graphicx}
\usepackage{amsmath}
\usepackage{multirow}
\usepackage{float}
\usepackage{amssymb}
\usepackage{booktabs}
\usepackage{subcaption}

\newcommand{\cmark}{\checkmark}
\newcommand{\xmark}{\ding{55}}
\usepackage{xcolor}

\usepackage{pifont}

\begin{document}

% \title{From Local Matches to Global Masks: Novel Instance Detection in Open-World Scenes}

\title{From Local Matches to Global Masks: Template-Guided Instance Detection and Segmentation in Open-World Scenes}

% You will get a Paper-ID when submitting a pdf file to the conference system
% \author{Author Names Omitted for Anonymous Review. Paper-ID 470}

\author{
\IEEEauthorblockN{
Qifan Zhang, Sai Haneesh Allu, Jikai Wang, Yangxiao Lu, Yu Xiang
}
\IEEEauthorblockA{
Intelligent Robotics and Vision Lab, The University of Texas at Dallas\\
%Richardson, TX, USA\\
\{qifan.zhang, saihaneesh.allu, jikai.wang, yangxiao.lu, yu.xiang\}@utdallas.edu
}
}

%\author{\authorblockN{Michael Shell}
%\authorblockA{School of Electrical and\\Computer Engineering\\
%Georgia Institute of Technology\\
%Atlanta, Georgia 30332--0250\\
%Email: mshell@ece.gatech.edu}
%\and
%\authorblockA{Twentieth Century Fox\\
%\and
%\authorblockN{James Kirk\\ and Montgomery Scott}
%\authorblockA{Starfleet Academy\\
%San Francisco, California 96678-2391\\

% avoiding spaces at the end of the author lines is not a problem with
% conference papers because we don't use \thanks or \IEEEmembership

%Homer Simpson\authorrefmark{2},
%Montgomery Scott\authorrefmark{3} and
%Eldon Tyrell\authorrefmark{4}}
%Telephone: (800) 555--1212, Fax: (888) 555--1212}
%\authorblockA{\authorrefmark{4}Tyrell Inc., 123 Replicant Street, Los Angeles, California 90210--4321}}

\makeatletter
\let\@oldmaketitle\@maketitle
\renewcommand{\@maketitle}{\@oldmaketitle
\centering
\vspace{2em}
\includegraphics[width=0.95\linewidth, trim={1.1cm 0 0.8cm 0}]{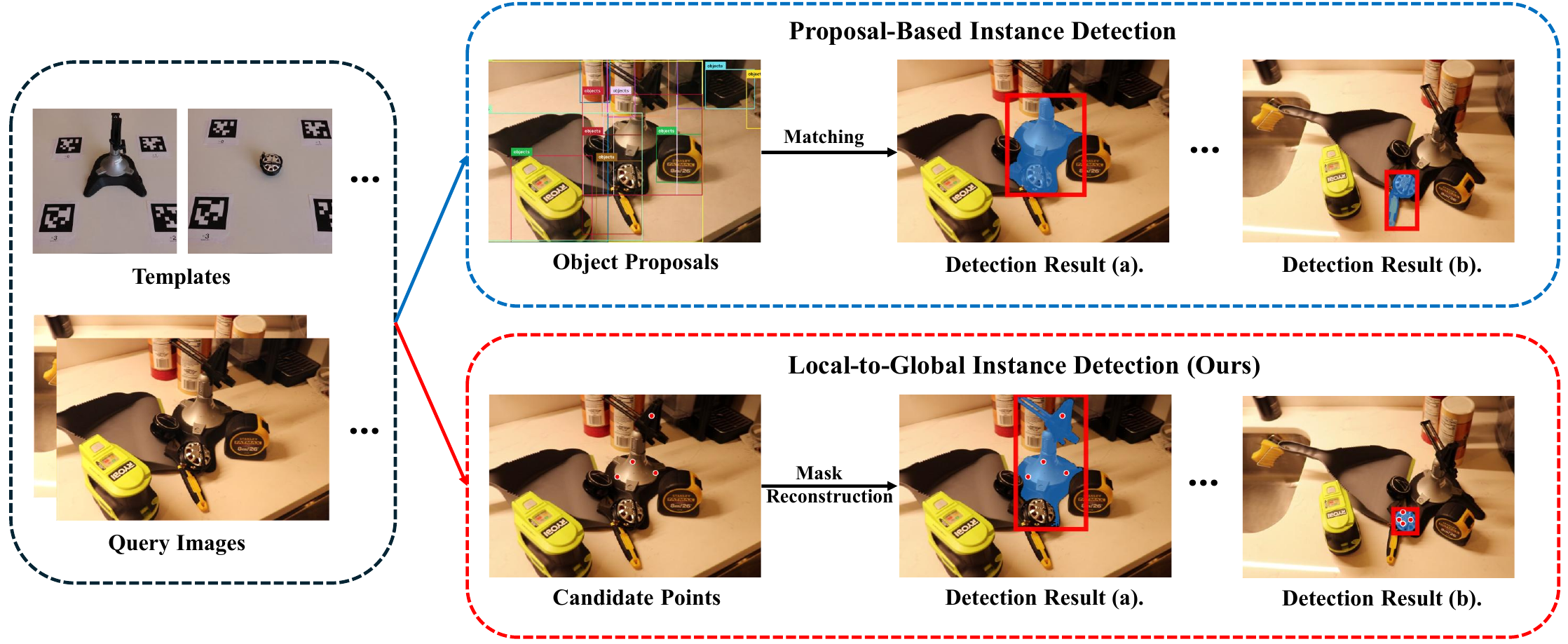}
% \vspace{-1.5em}
\captionof{figure}{
Conceptual comparison between object proposal-based instance detection methods and our local-to-global instance detection framework.
Top: Proposal-based approaches~\cite{lu2024adapting} first generate object proposals in the query image and then perform instance matching to obtain detection results.
Bottom: Our method starts from dense local correspondences to identify candidate points and reconstructs complete instance masks through mask reconstruction, producing final detection results without explicit proposal generation.
}\label{fig:intro}
%\vspace{-0.98em}
}
\makeatother

\maketitle

\begin{abstract}
Detecting and segmenting novel object instances in open-world environments is a fundamental problem in robotic perception. Given only a small set of template images, a robot must locate and segment a specific object instance in a cluttered, previously unseen scene. Existing proposal-based approaches are highly sensitive to proposal quality and often fail under occlusion and background clutter. We propose L2G-Det, a local-to-global instance detection framework that bypasses explicit object proposals by leveraging dense patch-level matching between templates and the query image. Locally matched patches generate candidate points, which are refined through a candidate selection module to suppress false positives. The filtered points are then used to prompt an augmented Segment Anything Model (SAM) with instance-specific object tokens, enabling reliable reconstruction of complete instance masks. Experiments demonstrate improved performance over proposal-based methods in challenging open-world settings. ~\footnote{Project website: \url{https://irvlutd.github.io/L2G/}}
\end{abstract}

\IEEEpeerreviewmaketitle

\setcounter{figure}{1} %

\section{Introduction}

%Object instance detection and segmentation in open-world environments~\cite{kim2022learning,nguyen2023cnos} has become increasingly important in robotic perception. In realistic settings, a system is often provided with only a small set of multi-view template images of a target object, and must localize this novel object within cluttered scenes while also producing an accurate instance segmentation mask. Such capabilities are essential for downstream manipulation, interaction, and scene understanding.

Object instance detection and segmentation in open-world environments~\cite{kim2022learning,nguyen2023cnos} have become increasingly important in robotic perception. In this problem setting, a robot is given a small set of template images of a target object, often captured from multiple viewpoints and is required to locate that specific object instance in a novel, cluttered scene. The goal is not only to determine whether the object is present, but also to precisely localize it and recover an accurate instance segmentation mask.

This capability is fundamental to many real-world robotic applications. For example, a service robot may be instructed to retrieve a particular household item shown to it only once, or a warehouse robot may need to pick a newly introduced product for which no prior training data exists. As robots increasingly operate in open-world environments, the ability to find and segment novel object instances from a small number of visual templates becomes a core perception capability.

% In manipulation tasks, accurate instance segmentation enables reasoning about object boundaries and visible surfaces for reliable grasp planning, while robust instance localization supports pose estimation and collision-aware motion execution in cluttered environments. More broadly, instance-level perception allows robots to track objects over time, distinguish between multiple similar instances, and ground task-level commands such as “pick up this object” in visual observations. 

Recent approaches~\cite{kim2022learning,li2024voxdet,lu2024adapting,shen2025solving} predominantly adopt an object proposal-based pipeline. As illustrated in Fig.~\ref{fig:intro}(top), an object proposal generator such as~\cite{kim2022learning,liu2023groundingDINO} is first applied to the query image to produce object-like regions, after which template embeddings are matched against these proposals to identify the target instance. However, the effectiveness of this pipeline is critically dependent on the quality of the proposal. Inaccurate proposals, such as regions covering only partial object parts or regions affected by background clutter, directly degrade the subsequent embedding matching stage and lead to degraded detection and segmentation performance. These limitations are particularly pronounced in real-world robotic environments, where objects frequently undergo occlusions, appear under diverse viewpoints, or are only partially visible. Under such conditions, proposal-based detectors struggle to produce high-quality proposals that align well with the complete object appearances provided by the templates, as illustrated in Fig.~\ref{fig:intro}(top).

To address these issues, we propose a new local-to-global instance detection framework (L2G-Det) centered on dense local feature matching. As illustrated in Fig.~\ref{fig:intro}(bottom), our approach first extracts dense patch-level features from template images using DINOv3~\cite{simeoni2025dinov3}, where each patch represents a localized object cue. For each template patch within the object, we identify its best-matching location in the query image and treat the center of the corresponding patch as a candidate point. By aggregating candidate points across multiple template views, we obtain a rich set of object-specific local cues. Instead of relying on explicit object proposals, we leverage these locally matched candidate points to reconstruct the global target object mask.

% \begin{figure*}[t]
%     \centering
%     \includegraphics[width=\linewidth]{intro_V3.pdf}
%     \caption{Conceptual comparison between proposal-based instance detection methods and our local-to-global instance detection framework.
% Top: Proposal-based approaches first generate object proposals in the query image and then perform instance matching to obtain detection results.
% Bottom: Our method starts from dense local correspondences to identify candidate points and reconstructs complete instance masks through mask reconstruction, producing final detection results without explicit proposal generation. \yu{Add some "three dots" to the templates and query images}}
%     \label{fig:intro}
% \end{figure*}

However, dense local matching alone inevitably introduces false positives due to local appearance ambiguities, where background regions or distractor objects share similar local textures or patterns with the target instance. To suppress such erroneous matches, we introduce a candidate selection module that refines candidate points. By probing each candidate point as a prompt with SAM~\cite{kirillov2023SAM} to obtain a local mask, and then comparing masked region embeddings against template embeddings, the candidate selector effectively filters unreliable matches while preserving locally consistent object cues. We also introduce a lightweight adapter~\cite{gao2021clipAdapter} to strengthen the ability of patch embeddings based on contrastive learning~\cite{tian2020makes}.

% As shown in Fig.~\ref{fig:SAM}, a pre-trained SAM model often focuses only on local regions around the prompt points and fails to recover the full object extent.

During mask generation, we use the filtered candidate points as prompts for SAM. However, these candidate points may not cover the entire object, which can result in incomplete masks. To address this limitation, we propose an \emph{Augmented SAM} module that incorporates a learnable, instance-specific object token~\cite{lester2021power, jia2022visual} into the mask decoder. This object token guides the frozen decoder to complete missing object parts and recover coherent global masks. 
The object tokens are learned using template-based synthetic images and stored in a memory pool, enabling incremental addition of instance-specific object tokens for new object instances without affecting previously learned ones.

In general, our approach achieves the state-of-the-art performance on two challenging instance detection benchmarks~\cite{shen2023instance,li2024voxdet} and demonstrates strong real-world performance in robotic experiments involving object search and navigation in cluttered environments.

% To further enhance robustness in complex environments, we introduce a \textbf{template-based learning} framework. Using only the template images, we composite the object onto diverse open-world backgrounds and perform few-shot fine-tuning in two complementary components:  
% (1) an adapter that strengthens the embedding’s ability to cluster views of the same object while separating different objects, and  
% (2) a prompt-based fine-tuning of SAM that improves its ability to reconstruct the object mask from sparse prompts.  
% For the latter, we freeze the SAM backbone and train a set of prompt tokens specific to each object. These prompts form a \emph{memory pool} that can be incrementally expanded as new objects are introduced, enabling continual learning and efficient adaptation without modifying the backbone.

Our main contributions are summarized as follows:
\begin{itemize}
    \item \textbf{Local-to-global novel instance detection.}
    We propose a local-to-global framework that bypasses object proposals by
    reconstructing global instance masks from dense local correspondences,
    enabling robust novel instance detection in cluttered scenes.
    % We propose a novel local-to-global framework for novel instance detection that outperforms proposal-based methods. Our method performs dense patch-level matching to identify locally consistent candidate points and progressively reconstructs global object masks from these local cues. 
    % This design enables robust instance detection under severe occlusion, viewpoint variation, and cluttered scenes.

     \item \textbf{Candidate selection via dense local matching.}
    We introduce a candidate selector that leverages multi-view templates to
    filter unreliable local matches and suppress false positives caused by
    appearance ambiguities.

    %  We introduce a candidate selection mechanism that leverages multiple template images to guide dense patch-level feature matching. 
    % By filtering unreliable matches and retaining locally consistent candidate points across templates, the proposed Candidate Selector effectively suppresses false positives caused by local appearance ambiguities, providing reliable prompts for downstream segmentation.

    \item \textbf{Template-based instance-specific object tokens.} 
    We propose an instance-specific object token memory that supports incremental learning of novel objects without
    interfering with previously learned instances.
    
    % We further propose an object token memory mechanism that stores instance-specific object tokens learned from template images. 
    % Each object is associated with an independent token, enabling incremental addition of new instances without affecting previously learned objects. 
    % This parameter-isolated design supports continual and scalable novel instance detection in open-world robotic environments.

    % \item \textbf{State-of-the-art performance and real-world robotic validation.}  
    % Our method achieves state-of-the-art results on multiple benchmark datasets and demonstrates strong real-world performance in robotic experiments, highlighting its capability to handle challenging novel instance detection tasks in practical scenarios.

    % We conduct comprehensive experiments on multiple benchmarks, demonstrating state-of-the-art performance compared to existing NID methods. 
    % In addition, we validate our approach on real-world robotic platforms, highlighting its practical effectiveness for novel instance detection and segmentation in realistic manipulation scenarios.
\end{itemize}

\begin{figure*}

  \centering
  \includegraphics[width=\linewidth,trim={0 0 10mm 0},clip]{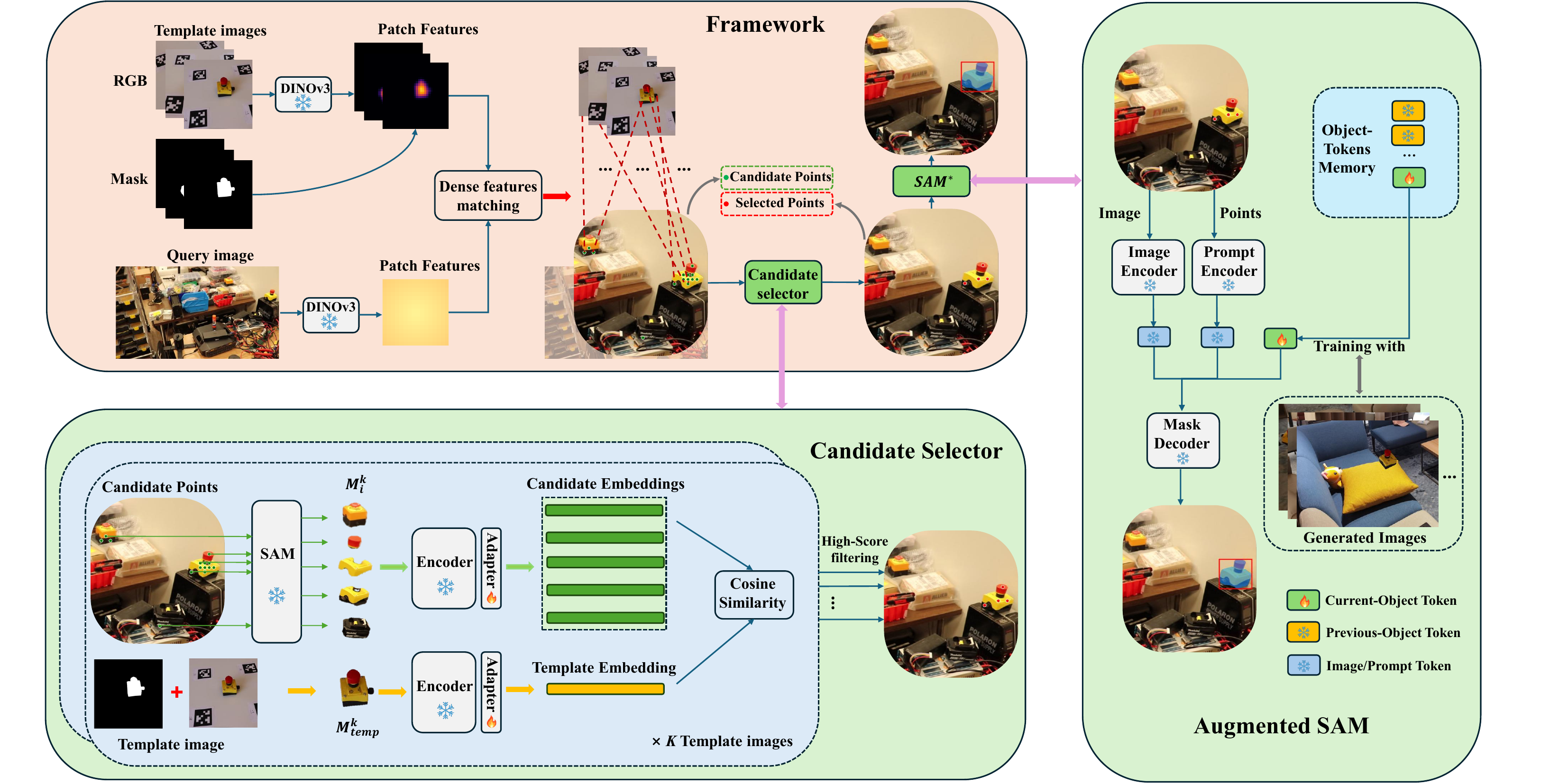}
  \caption{Overview of our L2G-Det framework for novel instance detection. It consists of a candidate selection module and an augmented SAM module (SAM$^{*}$). Only the adapters and object-tokens are learnable, while all other components are frozen.}
\vspace{-4mm}
  \label{fig:framework}
\end{figure*}

\section{Related Work}

\textbf{Instance Detection.}
Instance detection~\cite{bormann2021real, mercier2021deepTemplate,li2024voxdet, shen2025solving}, aims to localize and segment specific object instances
that are unseen during training, typically given only a small set of template
images. Most existing approaches adopt a proposal-based~\cite{lu2024adapting, shen2023instance} instance matching pipeline, where object-proposals are first generated and then matched against template embeddings. In addition to generating 2D representations~\cite{ammirato2018targetInstDet, mercier2021deepTemplate, liu2022gen6d}, VoxDet~\cite{li2024voxdet} further leverages multi-view observations
to construct 3D representations. While effective in relatively clean settings, these methods are highly sensitive
to proposal quality. In contrast, our method avoids explicit proposal generation and instead reconstructs global
object masks by aggregating locally matched features.

\textbf{Dense Local Feature Matching.}
Recent pretrained visual encoders~\cite{caron2021dino,bolya2025PerceptionEncoder} produce semantically meaningful patch-level features~\cite{dosovitskiy2020ViT} that generalize across objects and viewpoints, enabling dense matching between template and query images.
Several works exploit such correspondences for localization~\cite{sarlin2020superglue, sun2021loftr} or pose estimation~\cite{wang2019densefusion}. However, they typically focus on geometric alignment~\cite{wang2024eloftr} or sparse keypoints~\cite{lowe2004distinctive, bay2006surf} rather than instance-level segmentation.
Our approach builds upon dense patch-level matching~\cite{simeoni2025dinov3} but extends it to novel
instance detection by aggregating locally consistent matches across multiple
template views and reconstructing global object masks~\cite{kirillov2023SAM} from these local cues.

\textbf{Foundation Models.}
Foundation models refer to general-purpose visual models pretrained on large-scale
and diverse data, which provide highly transferable representations for a wide
range of downstream tasks.
Pretrained visual encoders~\cite{bolya2025PerceptionEncoder,caron2021dino,radford2021clip} have demonstrated strong capability in extracting semantically consistent and instance-discriminative features, making them well
suited for feature matching and instance-level perception. Proposal-based instance detection methods typically rely on detectors such as GroundingDINO~\cite{liu2023groundingDINO}, which extend foundation models to
open-vocabulary object localization by generating object proposals. In contrast, our method leverages Segment Anything (SAM)~\cite{kirillov2023SAM, ravi2024sam2, carion2025sam3segmentconcepts} to reconstruct instance
masks directly from locally matched features obtained via dense feature matching, without relying on proposal generation.
Unlike existing approaches~\cite{chen2023sam,yu2024ts} that fine-tune SAM for task-specific adaptation, we target the novel instance detection setting and introduce instance-specific object tokens to guide mask reconstruction, making our approach better suited for open-world, real-world environments.

\section{Method}

Most prior instance detection methods rely on an explicit object proposal detector. However, in cluttered open-world scenes, the quality of the proposal can be unreliable (partial boxes, background leakage, or missed instances), directly degrading template matching and final segmentation. To avoid this dependency, we adopt a local-to-global strategy that starts from dense local correspondences rather than global proposals (Fig.~\ref{fig:framework}). Specifically, we first perform dense patch-level matching between each template view and the query image to obtain candidate points that likely lie on the target instance. Since dense matching can introduce false positives due to local appearance ambiguities, we then design a Candidate Selector to filter candidate points. Finally, using the filtered points as prompts, we reconstruct the full-instance mask with an augmented SAM decoder that is guided by a learnable instance-specific object token. We describe these components in this section.

Our goal is to detect and segment a target object instance given a set of template images of that object.
We assume that each object instance is associated with $K$ template images
$\{\mathcal{I}_{\text{temp}}^{k}\}_{k=1}^{K}$ with corresponding object masks
$\{\mathcal{M}_{\text{temp}}^{k}\}_{k=1}^{K}$.
These template images characterize the appearance of the instance from multiple viewpoints. Given a query image $\mathcal{I}_{\text{query}}$, our objective is to localize the target instance and predict its segmentation mask by using the visual correspondence between the template images and the query image.

% The objective of our model is to detect and segment a novel instance given a set of template images. 
% We assume that for each of the \(N\) target instances, we are provided with \(K\) template images and their corresponding masks. 
% These template observations characterize the appearance of each instance from multiple viewpoints.
% We next present the details of our proposed framework.

\subsection{Dense Feature Matching}

\label{sec:dense_matching}

As illustrated in the \textbf{Framework} of Fig.~\ref{fig:framework}, we first perform dense feature matching between the template images and the query image to obtain an initial set of candidate points.
Our model employs a frozen DINOv3~\cite{simeoni2025dinov3} backbone to extract dense patch-level features.

For a given template image $\mathcal{I}_{\text{temp}}^{k}$, we uniformly sample $S$ patches inside the corresponding object mask $\mathcal{M}_{\text{temp}}^{k}$.
Let $\mathbf{f}_{i}^{k} \in \mathbb{R}^{D}$ denote the feature embedding of the $i$-th sampled patch from the $k$-th template image, where $i = 1, \dots, S$ and $D$ is the dimension of the feature embedding.
Similarly, we extract dense patch features
$\{\mathbf{f}_{j}\}_{j=1}^{N}$ from the query image $\mathcal{I}_{\text{query}}$ using the same DINOv3 encoder, where $N$ denotes the number of feature embeddings from the query image.

To match the feature embeddings, for each template patch feature $\mathbf{f}_{i}^{k}$, we compute its cosine similarity with all patch features $\{\mathbf{f}_{j}\}_{j=1}^{N}$ in the query image.
% \begin{equation}
% \label{eq:cosine_similarity}
% \mathrm{sim}\!\left(\mathbf{f}_{i}^{k}, \mathbf{f}_{j}\right)
% =
% \frac{
% \mathbf{f}_{i}^{k} \cdot \mathbf{f}_{j}
% }{
% \left\lVert \mathbf{f}_{i}^{k} \right\rVert_2
% \left\lVert \mathbf{f}_{j} \right\rVert_2
% },
% \end{equation}
Then for each template patch, we select the query patch with the highest cosine similarity score:
\begin{equation}
\label{eq:best_match}
j^{*}
=
\arg\max_{j \in \{1,\dots,N\}}
\mathrm{sim}\!\left(\mathbf{f}_{i}^{k}, \mathbf{f}_{j}\right).
\end{equation}

The spatial center of the selected query patch $j^{*}$ is then taken as a candidate point $\mathbf{p}_{i}^{k}$ corresponding to the $i$-th patch from the $k$-th template.
Repeating this process for all $S$ sampled patches yields a set of candidate points for the $k$-th template:
\begin{equation}
\label{eq:candidate_points}
\mathcal{C}^{k}
=
\left\{
\mathbf{p}_{i}^{k}
\;\middle|\;
i = 1, \dots, S
\right\}.
\end{equation}
Finally, we apply this procedure independently to each template image to generate $K$ candidate point sets
$\{\mathcal{C}^{k}\}_{k=1}^{K}$, which serve as the initial candidate points for subsequent refinement and selection.

% Our model employs DINOv3 to extract dense patch-level features. 
% Given a template image of the target object, we uniformly sample \(S\) patches inside the object mask, 
% referred to as \textit{mask patches}. For each mask patch, we compute its feature vector and match it 
% against all patch features in the query image using cosine similarity. The location of the highest-scoring 
% patch in the query image is selected as a candidate point. After deduplication and filtering, the set of 
% candidate points obtained from the \(m\)-th template image is denoted as:
% \[
% C_m = \{ p_1, p_2, \ldots, p_i \}.
% \]

% Aggregating candidates across all template images, the full candidate set for the object is:
% \[
% C_{\text{all}} = C_1 \cup C_2 \cup C_3 \cup \cdots \cup C_m.
% \]
% Here, \(C_{\text{all}}\) represents the union of all candidate points collected from the \(m\) available 
% template images of the current object.

\subsection{Candidate Selector}

\label{sec:candidate_selector}

After performing dense feature matching using patch features extracted by DINOv3, we obtain an initial set of candidate points.
However, many of these candidates do not fall within the target object region and are falsely matched.
This is mainly caused by local regions in the query image that exhibit similar appearance to certain local parts of the target instance.
As illustrated in the \textbf{Framework} of Fig.~\ref{fig:framework}, red dashed lines indicate dense feature correspondences, while the green points denote the initial candidate points, among which some lie outside the true object region.
Therefore, an additional filtering step is required to refine the candidate set.

The \textbf{Candidate Selector} in Fig.~\ref{fig:framework} depicts the filtering process applied in a template-wise manner.
For each template image, we perform the same candidate selection pipeline independently, and finally aggregate the selected points from all templates, corresponding to the red \emph{Selected Points} shown in the figure.

\paragraph{Single-point SAM probing}
Given a template image $\mathcal{I}_{\text{temp}}^{k}$, we first consider its associated candidate point set $\mathcal{C}^{k} = \{\mathbf{p}_i^{k}\}_{i=1}^{S}$ obtained from dense feature matching.
Each candidate point $\mathbf{p}_i^{k}$ is individually used as a point prompt for the SAM model, which produces a corresponding mask region $\mathcal{M}_{i}^{k}, i=1,\ldots,S$ in the query image.
Since only a single point is provided as the prompt, the resulting mask typically focuses on a local region around the queried point, serving as a local feature probe.

\paragraph{Candidate and template embeddings}
The generated local mask $\mathcal{M}_{i}^{k}$ is applied to the query image to extract a masked local region, which is then processed by a frozen feature encoder $E(\cdot)$ followed by a learnable adapter to obtain the candidate embedding. Similarly, using the template image together with its object mask, we extract the full object region and compute the corresponding template embedding:
\begin{equation}
\label{eq:candidate_embedding}
\mathbf{z}_{i}^{k}
=
\mathcal{A}\!\left(
E\!\left(\mathcal{I}_{\text{query}} \odot \mathcal{M}_{i}^{k}\right)
\right),
\end{equation}
\begin{equation}
\label{eq:template_embedding}
\mathbf{z}_{\text{temp}}^{k}
=
\mathcal{A}\!\left(
E\!\left(\mathcal{I}_{\text{temp}}^{k} \odot \mathcal{M}_{\text{temp}}^{k}\right)
\right),
\end{equation}
where $\odot$ denotes element-wise masking, $E(\cdot)$ is a pretrained encoder for visual feature extraction, and $\mathcal{A}(\cdot)$ is a lightweight residual MLP adapter.

\paragraph{Learnable residual MLP adapter}
The adapter $\mathcal{A}(\cdot)$ is implemented as a residual MLP:
\begin{equation}
\label{eq:adapter}
\mathcal{A}(\mathbf{x})
=
\mathbf{x}
+
\alpha \cdot \mathrm{MLP}\!\left(\mathbf{x}\right),
\end{equation}
where $\mathbf{x}$ denotes the input to the adapter, $\mathrm{MLP}(\cdot)$ consists of two linear layers with a non-linear activation in between, and $\alpha$ is a scaling factor.
All parameters of the backbone encoder $E(\cdot)$ are frozen, and only the adapter parameters are learnable.

% \paragraph{Similarity-based filtering.}
% To measure the relevance of each candidate point to the target instance, we compute the cosine similarity between the candidate embedding and the template embedding. Candidates are ranked according to their similarity scores, and high-confidence candidates are retained for subsequent processing.
% The selected candidates from all template images are then aggregated to form the final set of selected points.
% % \begin{equation} 
% % \label{eq:candidate_similarity}
% % s_{i}^{k}
% % =
% % \frac{
% % \mathbf{z}_{i}^{k} \cdot \mathbf{z}_{t}^{k}
% % }{
% % \lVert \mathbf{z}_{i}^{k} \rVert_2
% % \lVert \mathbf{z}_{t}^{k} \rVert_2
% % }.
% % \end{equation}
The adapter is introduced to further enhance instance-level discrimination via contrastive learning on the given template images.
Given embeddings from the same object instance as positives and embeddings from different instances as negatives, we employ an InfoNCE-style contrastive loss~\cite{oord2018representation, chen2020simple} to learn the adapter:
\begin{equation}
\label{eq:contrastive_loss}
\mathcal{L}_{\text{con}}
=
-
\log
\frac{
e^{\mathrm{sim}(\mathbf{z}, \mathbf{z}^{+}) / \tau}
}{
e^{\mathrm{sim}(\mathbf{z}, \mathbf{z}^{+}) / \tau}
+
\sum\limits_{\mathbf{z}^{-}}
e^{\mathrm{sim}(\mathbf{z}, \mathbf{z}^{-}) / \tau}
},
\end{equation}
where $\mathbf{z}$ and $\mathbf{z}^{+}$ denote embeddings from the same object instance, $\mathbf{z}^{-}$ denotes embeddings from different instances, $\tau$ is a temperature parameter, and $\mathrm{sim}(\cdot,\cdot)$ denotes cosine similarity. We freeze the visual encoder and train the learnable adapter using the given template images and template-based synthetic images, as described in Sec.~\ref{para:template_synthetic}.

After obtaining the candidate embeddings and the corresponding template embedding, we compute the cosine similarity between each candidate embedding and the template embedding from Eq.~\eqref{eq:candidate_embedding} and Eq.~\eqref{eq:template_embedding}:
\begin{equation}
\label{eq:matching_score}
s_i^k
=
\frac{\mathbf{z}_i^k \cdot \mathbf{z}_{\mathrm{temp}}^k}
{\lVert \mathbf{z}_i^k \rVert_2 \, \lVert \mathbf{z}_{\mathrm{temp}}^k \rVert_2},
\end{equation}
where $s_i^{k}$ denotes the similarity score between the $i$-th candidate embedding and the template embedding for the $k$-th template image.
The candidates are then ranked according to their similarity scores. We first identify the candidate point with the maximum similarity score: $s_{\max}^{k} = \max_{i} \; s_i^{k}$. We then apply a filtering threshold $\delta$ and retain not only the candidate point with the highest score, but also all candidate points whose similarity scores differ from the maximum by less than $\delta$:
\begin{equation}
\label{eq:threshold_filtering}
\mathcal{P}^{k}
=
\left\{
\mathbf{p}_i^{k}
\;\middle|\;
s_{\max}^{k} - s_i^{k} < \delta
\right\},
\end{equation}
where $\mathcal{P}^k$ denotes the set of selected candidate points retained for the $k$-th template image after filtering. This strategy ensures that the most similar regions to the target instance are preserved, while also retaining secondary yet relevant local regions.
Such regions often correspond to meaningful object parts (e.g., the head of the target object in the local mask $\mathcal{M}_i^{k}$ shown in the Candidate Selector of Fig.~\ref{fig:framework}), which also exhibit high similarity scores.
At the same time, low-confidence candidates originating from non-target regions are effectively filtered out. The above procedure is applied independently to all $K$ template images.
Finally, the selected points from all templates are aggregated to form the final set of selected candidate points: $
\mathcal{P}
=
\bigcup_{k=1}^{K}
\mathcal{P}^{k}$.

\subsection{Augmented SAM}
\label{sec:augmented_sam}

\begin{figure}

  \centering
  \includegraphics[width=0.8\linewidth]{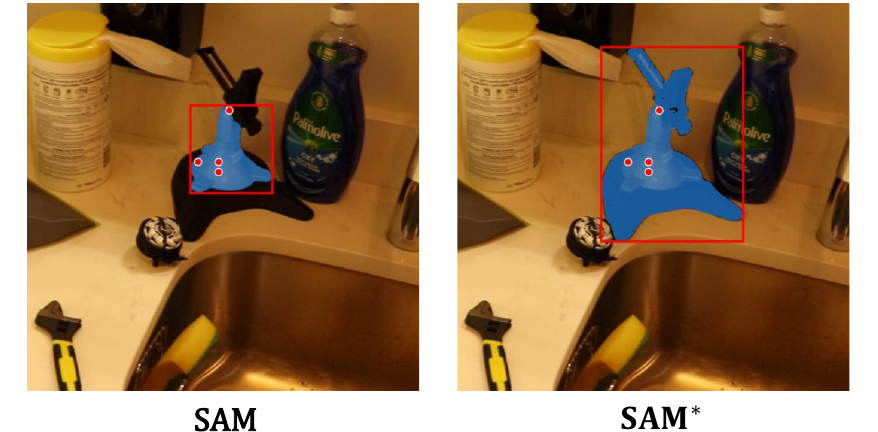}
  \caption{Comparison between the original SAM and our augmented \textbf{SAM}$^{*}$ under candidate points. With all parameters frozen, SAM tends to produce incomplete masks focused on local regions around the prompts. In contrast, \textbf{SAM}$^{*}$ incorporates a learnable instance-specific object token, which guides the decoder to produce more complete object masks improving detection performance. }
\vspace{-4mm}
  \label{fig:SAM}
\end{figure}

The selected points $\mathcal{P}$ obtained from the Candidate Selector (red points in Fig.~\ref{fig:framework}) are designed to lie on the target instance as much as possible.
However, sparse point prompts cannot guarantee coverage of all informative object parts.
As illustrated in Fig.~\ref{fig:SAM}, the selected points $\mathcal{P}$ may concentrate on the central region of the target instance, while other local regions remain uncovered.
This phenomenon arises because certain object parts exhibit local features that differ significantly from the overall appearance of the instance, causing the corresponding candidate points to be filtered out by the \emph{Candidate Selector}. As a result, using these selected points with frozen SAM typically leads to partial masks that cover only local regions of the target object.
Therefore, we seek to further enhance SAM’s ability to reconstruct complete and accurate masks to address this limitation.

\paragraph{Object token for mask completion}
To this end, we introduce a learnable \emph{object token} whose purpose is to guide SAM toward predicting the full object mask.
The object token is injected into the SAM mask decoder together with the image tokens and prompt tokens, as illustrated in the \textbf{Augmented SAM} module of Fig.~\ref{fig:framework}.
The image encoder and prompt encoder are kept frozen, and therefore the corresponding image tokens and prompt tokens are also fixed during training.

Formally, given the query image $\mathcal{I}$ and a set of prompt points $\mathcal{P}$ after the \emph{Candidate Selector}, the image tokens and prompt tokens are obtained as
\begin{equation}
\label{eq:image_prompt_tokens}
\mathbf{T}_I = E_I(\mathcal{I}), \qquad
\mathbf{T}_P = E_P(\mathcal{P}),
\end{equation}
where $E_I(\cdot)$ and $E_P(\cdot)$ denote the frozen SAM image encoder and prompt encoder, respectively.
For each object instance, we introduce an instance-specific learnable object token $\mathbf{t}_O$.
The predicted mask is then produced by the frozen mask decoder conditioned on all three types of tokens:
\begin{equation}
\label{eq:augmented_sam_decoder}
\hat{\mathcal{M}} = D_{\text{mask}}\!\left(\mathbf{T}_I, \mathbf{T}_P, \mathbf{t}_O \right),
\end{equation}
where $D_{\text{mask}}(\cdot)$ denotes the SAM mask decoder.
During training, only the object token $\mathbf{t}_O$ is learnable, while all other components remain frozen.

% \paragraph{Template-based synthetic training data.}
% \label{para:template_synthetic}
% For each instance, we train its corresponding object token using the given template images.
% Specifically, we composite the masked object regions from the template images onto publicly available open-world background images.
% We consider three types of synthesized scenes:
% \begin{enumerate}
%     \item \textbf{Single-object composition:} only the target object is pasted onto the background image.
%     \item \textbf{Multi-object composition without overlap:} the target object and several other objects are pasted onto the background, with additional objects placed around the target object without overlapping it.
%     \item \textbf{Multi-object composition with overlap:} both the target object and other objects are pasted onto the background, where objects may partially overlap.
%     When the target object is in the frontmost position, it remains unoccluded; otherwise, it may be partially occluded by surrounding objects, simulating challenging real-world conditions.
% \end{enumerate}

\paragraph{Template-based synthetic training data}  \label{para:template_synthetic}

Based on the given template images and their corresponding mask images, we generate a set of synthetic training data to train both the adapter in the Candidate Selector and the object token in the Augmented SAM.
Specifically, we composite the masked object regions from the template images onto publicly available open-world background images, which are provided
in~\cite{shen2023instance}. We consider three types of synthesized scenes:

% \begin{figure}

%   \centering
%   \includegraphics[width=1\linewidth]{Syn_imgs.pdf}
%   \caption{Examples of template-based synthetic training images, where the target object is highlighted by the red dashed outline. \yu{show images for each type. So the text can mention these images.}}
% \vspace{-1mm}
%   \label{fig:Syn}
% \end{figure}

% \yu{show a figure with these images}

% We consider three types of synthesized scenes: \textbf{1) Single-object composition:} only the target object is pasted onto the background image; \textbf{2) Multi-object composition without overlap:} the target object and several other objects are pasted onto the background, with additional objects placed around the target object without overlapping it; \textbf{3) Multi-object composition with overlap:} both the target object and other objects are pasted onto the background, where objects may partially overlap.
%     When the target object is in the frontmost position, it remains unoccluded; otherwise, it may be partially occluded by surrounding objects, simulating challenging real-world conditions.

\begin{enumerate}
    \item \textbf{Single-object composition:} only the target object is pasted onto the background image.
    \item \textbf{Multi-object composition without overlap:} the target object and several other objects are pasted onto the background, with additional objects placed around the target object without overlapping it.
    \item \textbf{Multi-object composition with overlap:} both the target object and other objects are pasted onto the background, where objects may partially overlap.
    When the target object is in the frontmost position, it remains unoccluded; otherwise, it may be partially occluded by surrounding objects, simulating challenging real-world conditions.
\end{enumerate}

These template-based synthetic images (check details in Appendix~\ref{Appendix:Sync_images}) enrich the training data and facilitate effective learning of both the adapter in the Candidate Selector and the object token in the Augmented SAM.
Moreover, this template-based compositing strategy significantly reduces the computational cost and time required for data generation.
Experimental results further demonstrate that such a simple synthesis approach can effectively improve model performance, without the need for additional overhead from complex generative models.
This simplicity makes the proposed method particularly suitable for novel instance recognition in real-world environments.

\paragraph{Training Loss}
Given a synthesized training image $\tilde{\mathcal{I}}$ and its ground-truth target mask $\tilde{\mathcal{M}}$, we randomly sample between $1$ and $m$ pixel locations within the target object mask.
These sampled points are constrained to lie inside the object mask and are used as point prompts for SAM.
The ground-truth object mask $\tilde{\mathcal{M}}$ serves as the supervision signal.

The object token is optimized using a hybrid segmentation loss composed of the binary cross-entropy loss~\cite{goodfellow2016deep}, the Dice loss~\cite{milletari2016vnet}, and the IoU loss~\cite{rahman2016iou}:
% \begin{equation}
% \label{eq:segmentation_loss}
% \mathcal{L}_{\text{seg}}
% =
% \mathcal{L}_{\text{BCE}}
% +
% \lambda_{\text{Dice}}\, \mathcal{L}_{\text{Dice}}
% +
% \lambda_{\text{IoU}}\, \mathcal{L}_{\text{IoU}},
% \end{equation}
\begin{align}
\mathcal{L}_{\text{seg}}
&=
\mathcal{L}_{\text{BCE}}
+
\lambda_{\text{Dice}}\, \mathcal{L}_{\text{Dice}}
+
\lambda_{\text{IoU}}\, \mathcal{L}_{\text{IoU}}, \label{eq:segmentation_loss} \\
% \mathcal{L}_{\text{BCE}} &= \text{BCEWithLogits}(\hat{\mathcal{M}}, \tilde{\mathcal{M}}), \\
\mathcal{L}_{\text{Dice}} &= 1 - \frac{2|\hat{\mathcal{M}} \cap \tilde{\mathcal{M}}| + \epsilon}
{|\hat{\mathcal{M}}| + |\tilde{\mathcal{M}}| + \epsilon}, \\
\mathcal{L}_{\text{IoU}} &= 1 - \frac{|\hat{\mathcal{M}} \cap \tilde{\mathcal{M}}| + \epsilon}
{|\hat{\mathcal{M}} \cup \tilde{\mathcal{M}}| + \epsilon}.
\end{align}
In the above equations, $\lambda_{\text{Dice}}$ and $\lambda_{\text{IoU}}$ denote the weighting coefficients for the Dice loss and the IoU loss, respectively.
Here, $\tilde{\mathcal{M}}$ represents the ground-truth mask, and $\hat{\mathcal{M}}$ denotes the predicted mask.

% \paragraph{Object token memory for incremental instance addition.}
% We further introduce an \emph{object token memory} to support continual and scalable learning.
% Each learned object token is stored in the memory after training.
% As new object instances are introduced, new object tokens are appended to the memory, allowing the memory size to grow dynamically.
% Importantly, training a new object token does not modify previously stored tokens, preventing catastrophic forgetting of earlier instances.
% This design enables continual learning and improves scalability when adapting to new objects or object sets.

\paragraph{Object token memory for incremental instance addition}
To better address the challenge of novel instance detection in a template-based setting, we design an improved training strategy for SAM by introducing an object token memory that supports incremental and scalable instance addition.
Each learned object token corresponds to a specific object instance and is stored in the memory after training.
During inference, each object is associated with its own dedicated object token, ensuring that detecting or segmenting one object does not interfere with other instances. Since object tokens are instance-specific, the appropriate object token at inference time is directly determined by the provided template images of the target instance.

As new object instances are introduced, new object tokens are appended to the memory, allowing the memory size to grow dynamically.
Importantly, training a new object token does not modify previously stored tokens, thereby preventing catastrophic forgetting~\cite{kirkpatrick2017overcoming} of earlier instances.
This parameter-isolated design enables continual learning~\cite{li2017learning} and provides a scalable framework for incorporating new novel instances over time, which is particularly beneficial for long-term deployment in open-world environments.

\section{EXPERIMENTS}

\subsection{Datasets and Settings}

\textbf{Datasets.} We evaluate our method on two datasets designed for novel instance detection.
The High-Resolution Instance Detection (HR-InsDet) dataset~\cite{shen2023instance} contains 100 object instances and a total of 160 test images captured across 14 different indoor scenes.
Each test image has a high resolution of $8192 \times 6144$. The test images are further categorized into \emph{easy} and \emph{hard} subsets according to object clutter and scene complexity. We follow the experimental settings in~\cite{shen2023instance}. 
Specifically, FasterRCNN~\cite{ren2015faster}, RetinaNet~\cite{lin2017focal}, CenterNet~\cite{zhou2019objects}, FCOS~\cite{9010746}, and the transformer-based DINO~\cite{zhang2022dinodetr} detector are trained using synthetic images generated by the Cut-Paste-Learn ~\cite{dwibedi2017cut} strategy.
This approach constructs training data by pasting object instances onto background images, enabling supervised training for instance detection.
For non-learned methods, we directly employ \textbf{SAM}~\cite{kirillov2023SAM} and DINO~\cite{caron2021dino, oquab2023dinov2} to generate object proposals and visual features.
In addition, we include NIDS-Net~\cite{lu2024adapting}, a recent state-of-the-art method, which introduces a trainable adapter to enhance performance.

The RoboTools dataset~\cite{li2024voxdet} consists of 20 object instances evaluated in 24 complex scenes, with each test image having a resolution of $1920 \times 1080$. We evaluate methods with different proposal detectors. Specifically, OLN$_{\text{Corr.}}$ employs its own detection module~\cite{kim2022learning}, where a matching head based on correlation~\cite{liu2022gen6d}.
Several other methods utilize GroundingDINO as the proposal model.

\textbf{Evaluation metrics.}
We adopt \textbf{Average Precision (AP)} as the primary evaluation metric for instance detection.
AP is computed by averaging precision over multiple Intersection over Union (IoU) thresholds ranging from 0.50 to 0.95 with a step size of 0.05.
We additionally report AP$_{50}$ and AP$_{75}$, which correspond to IoU thresholds of 0.50 and 0.75, respectively.

\textbf{Implementation details.}
All experiments are conducted using two NVIDIA A5000 GPUs.
In the Candidate Selector, the residual adapter ratio $\alpha$ in Eq.~\eqref{eq:adapter} is set to 0.2. The filtering threshold $\delta$ in Eq.~\eqref{eq:threshold_filtering} is set to $0.01$.
We employ the Adam optimizer~\cite{kingma2014adam} with a learning rate of $5 \times 10^{-4}$, a batch size of 96, and train the adapter for 20 epochs.
For the InfoNCE-style contrastive learning~\cite{oord2018representation} objective, the temperature parameter $\tau$ is set to 0.07, and the ratio between positive and negative samples is 1:2. In the Augmented SAM module, all components of SAM are frozen except for the instance-specific object tokens. In the training loss Eq.~\eqref{eq:segmentation_loss}, we set $\lambda_{\text{Dice}} = 0.5$ and $\lambda_{\text{IoU}} = 1.0$, respectively.
We use the Adam optimizer with a learning rate of $5 \times 10^{-3}$, a batch size of 1, and train for 12 epochs.

For template-based synthetic training images, three types of synthesized scenes (Sec.~\ref{para:template_synthetic}) are used with an equal ratio of 1:1:1.
For data augmentation, we apply random scaling, random rotation, and random blur to the target object.
The background images are sampled from the open-world background dataset provided in~\cite{shen2023instance}. Ultimately, we generate 500 synthesized images for each target object. 
% Ultimately, we generate 500 synthesized images(examples in Fig.~\ref{fig:Syn}) for each target object.

% For dense feature extraction, we employ DINOv3-large~\cite{simeoni2025dinov3}.
% All SAM components are built upon SAM2-large~\cite{ravi2024sam2}.
% Within the Candidate Selector, a pre-trained Perception Encoder (PE)~\cite{bolya2025PerceptionEncoder} is used to extract visual embeddings for both candidate regions and template objects. 
% For dense feature matching, we uniformly sample $S=10$ patches inside the object mask of each template image.
% For the RoboTools benchmark and real-world robotic experiments, we directly perform inference on the full image.
% For the HR-InsDet benchmark, due to its high image resolution, we adopt a local-window inference strategy.
% Each local window has a resolution of $2048 \times 1536$, corresponding to one quarter of the original image resolution along each spatial dimension.

For dense feature extraction, we employ DINOv3-large~\cite{simeoni2025dinov3}.
All SAM components are built upon SAM2-large~\cite{ravi2024sam2}.
Within the Candidate Selector, a pre-trained Perception Encoder (PE)~\cite{bolya2025PerceptionEncoder} is used to extract visual embeddings for both candidate regions and template objects.
For dense feature matching, we uniformly sample $S=10$ patches inside the object mask of each template image.
After the Candidate Selector obtains the selected candidates, we further choose the final prompt points using a farthest-point-first strategy, which prevents the prompts from being concentrated in a small local region and encourages spatial diversity for mask reconstruction.
For the RoboTools benchmark and real-world robotic experiments, we directly perform inference on the full image.
For the HR-InsDet benchmark, due to its high image resolution, we adopt a local-window inference strategy.
Each local window has a resolution of $2048 \times 1536$, corresponding to one quarter of the original image resolution along each spatial dimension.

\subsection{Benchmarking Results}

\textbf{HR-InsDet.}
Table~\ref{tab:HR-InsDet} summarizes the results on the High-Resolution dataset.
Our L2G-Det achieves a substantial improvement of \textbf{12.3 AP} over the top-performing baseline.
Notably, on the hard subset, which contains severe occlusion and heavy clutter, L2G-Det yields an even larger gain of \textbf{17.6 AP}. These results demonstrate the effectiveness and robustness of our approach in complex real-world scenarios.

\begin{figure}

  \centering
  \includegraphics[width=\linewidth,trim={0 0 0 1.3cm},clip]{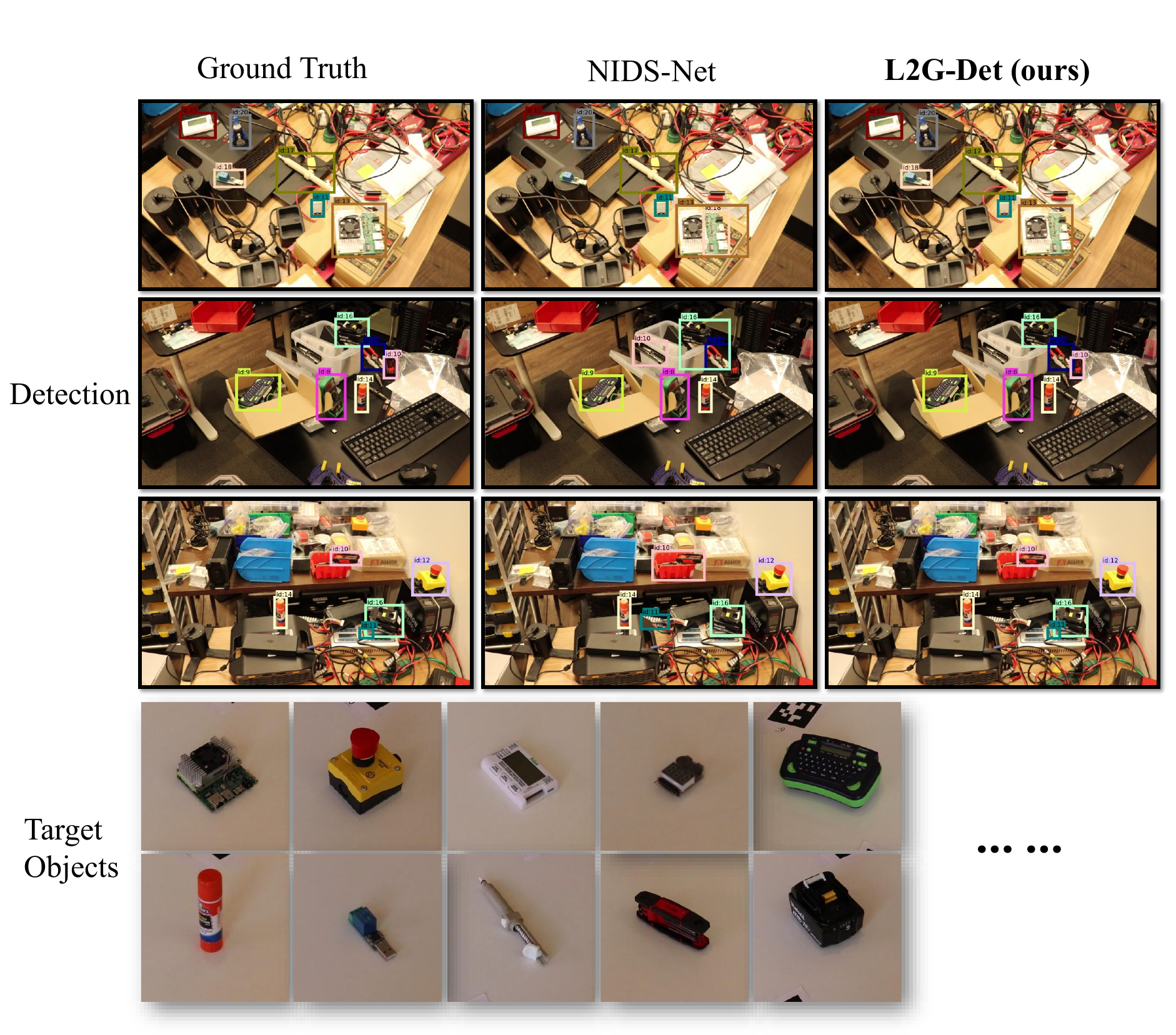}
  \caption{Qualitative results on RoboTools benchmark.
From left to right, we show the ground-truth annotations, results produced by NIDS-Net~\cite{lu2024adapting}, and results of our method \textbf{L2G-Det}.}
\vspace{-4mm}
  \label{fig:RoboTools}
\end{figure}

\textbf{RoboTools.}
Table~\ref{tab:RoboTools} reports the detection performance on the RoboTools dataset.
Unlike proposal-based pipelines, our L2G-Det does not rely on explicit proposal generation; instead, it performs dense patch-level matching to obtain candidate points for subsequent mask reconstruction.
As a result, L2G-Det outperforms the top-performing baseline, NIDS-Net~\cite{lu2024adapting}, by \textbf{7.0 AP}.
Qualitative comparisons on RoboTools are shown in Fig.~\ref{fig:RoboTools}, where our method produces more complete and accurate detections in cluttered scenes. More results can be found in the Appendix~\ref{Appendix:results}.

\begin{table}
\centering
\caption{Comparison of detection performance on the HR-InsDet dataset~\cite{shen2023instance}.}
\label{tab:HR-InsDet}
\scalebox{0.75}{
\begin{tabular}{lccccccccc}
\toprule
\multirow{2}{*}{Method} 
& \multicolumn{6}{c}{AP} 
& \multirow{2}{*}{AP$_{50}$} 
& \multirow{2}{*}{AP$_{75}$} \\
\cmidrule(lr){2-7}
& avg & hard & easy & small & medium & large & & \\
\midrule
FasterRCNN~\cite{ren2015faster}          & 19.5 & 10.3 & 23.8 &  5.0 & 22.2 & 38.0 & 29.2 & 23.3 \\
RetinaNet~\cite{lin2017focal}           & 22.2 & 14.9 & 26.5 &  5.5 & 25.8 & 42.7 & 31.2 & 25.0 \\
CenterNet~\cite{zhou2019objects}           & 21.1 & 11.8 & 25.7 &  5.9 & 24.1 & 40.4 & 32.7 & 23.6 \\
FCOS~\cite{9010746}                & 22.4 & 13.2 & 28.7 &  6.2 & 26.5 & 38.1 & 32.8 & 25.5 \\
DINO~\cite{zhang2022dinodetr}                & 28.0 & 17.9 & 32.6 & 11.5 & 31.6 & 48.3 & 39.6 & 32.2 \\
SAM + DINO$_f$~\cite{kirillov2023SAM,caron2021dino}      & 37.0 & 22.4 & 43.9 & 11.9 & 40.9 & 62.7 & 44.1 & 40.4 \\
SAM + DINOv2$_f$~\cite{kirillov2023SAM,oquab2023dinov2}     & 41.6 & 28.0 & 47.6 & 14.6 & 45.8 & 69.1 & 49.1 & 46.0 \\
% IDOW$_{\text{GroundingDINO}}$  & 57.0 & 40.7 & 64.4 & 35.3 & 63.0 & 73.6 & 69.3 & 62.8 \\
NIDS-Net~\cite{lu2024adapting}             & 63.9 & 43.4 & 72.7 &  18.1 & 62.5 &  84.0 & 76.6 & 70.6 \\
\midrule

\textbf{L2G-Det (Ours)}    & \textbf{76.2} & \textbf{61.0} & \textbf{81.2} & \textbf{24.1} & \textbf{75.4} & \textbf{92.6} 
                                 & \textbf{80.6} & \textbf{77.9} \\
\bottomrule
\end{tabular}
}
\vspace{-1mm}
\end{table}
% \vspace{-10pt}

\begin{table}
\centering
\caption{Comparison of detection performance on RoboTools dataset~\cite{li2024voxdet}. The \textit{Proposal} column indicates the proposal generation strategy. Specifically, OLN denotes Object Localization Network~\cite{kim2022learning}, GD denotes GroundingDINO~\cite{liu2023groundingDINO}.}
\label{tab:RoboTools}
\scalebox{0.8}{
\begin{tabular}{l@{\hspace{37pt}}c@{\hspace{37pt}}c@{\hspace{37pt}}c@{\hspace{37pt}}c}
\toprule
Method & Proposal& AP & AP$_{50}$ & AP$_{75}$ \\
\midrule
OS2D~\cite{osokin2020os2d}       &    N/A     & 2.9  & 6.5  & 2.0  \\
DTOID~\cite{mercier2021deepTemplate}       &    N/A     & 3.6  & 9.0  & 2.0  \\
OLN$_{\text{Corr.}}$~\cite{kim2022learning}   & OLN  & 14.4 & 18.1 & 15.7 \\
VoxDet~\cite{li2024voxdet}        &   OLN     & 18.7 & 23.6 & 20.5 \\
OTS-FM~\cite{shen2025solving, shen2023instance}        &   GD     & 56.7 & 64.8 & 59.0 \\
IDOW~\cite{shen2025solving}          &   GD   & 59.4 & 67.8 & 61.8 \\
NIDS-Net~\cite{lu2024adapting}       &  GD    & 64.9 & 79.4 & 70.8 \\
\midrule

\textbf{L2G-Det (Ours)} & \textbf{N/A}  & \textbf{71.9} & \textbf{84.6} & \textbf{77.2} \\
\bottomrule
\end{tabular}
}
\vspace{-4mm}
\end{table}

\subsection{Ablation Studies}

We analyze the importance and effectiveness of each component in our model through a series of ablation studies.
Unless otherwise specified or explicitly varied in a particular experiment, all ablation experiments follow the same settings described in the implementation details.

\textbf{Effect of the Adapter and Augmented SAM.}
Across both datasets, we observe that enabling either the adapter or Augmented SAM (i.e., introducing instance-specific object tokens) individually leads to consistent improvements over the base L2G-Det framework. As summarized in Table.~\ref{tab:ablation}, when both components are jointly applied, the detection performance is further enhanced, achieving the best results. Notably, the performance gains are more pronounced on the High-resolution dataset. This can be attributed to the presence of many visually similar objects, as well as increased object overlap and occlusion in query images.
In such challenging scenarios, the adapter improves instance-level feature discrimination during candidate selection, while Augmented SAM effectively recovers complete object masks from sparse prompts.
Overall, these results demonstrate the effectiveness and complementarity of the two components in improving the detection capability of L2G-Det.

\begin{table}[t]
\centering
\caption{Ablation study on the effects of the Adapter in Candidate Selector and Augmented SAM on RoboTools~\cite{li2024voxdet} and High-resolution datasets~\cite{shen2023instance}.}
\label{tab:ablation}
\resizebox{0.9\linewidth}{!}{
\begin{tabular}{lccc}
\toprule
\textbf{Method} & \textbf{Adapter} & \textbf{Augmented SAM} & \textbf{AP} \\
\midrule
\multicolumn{4}{l}{\textbf{RoboTools Dataset}} \\
\midrule
w/o Adapter + SAM                 & \xmark & \xmark & 64.8 \\
Adapter + SAM                & \cmark & \xmark & 67.5 \\
w/o Adapter + SAM$^{*}$             & \xmark & \cmark & 69 \\
Adapter + SAM$^{*}$               & \cmark & \cmark & \textbf{71.9} \\
\midrule
\multicolumn{4}{l}{\textbf{High-resolution Dataset}} \\
\midrule
w/o Adapter + SAM                  & \xmark & \xmark & 58.9 \\
Adapter + SAM                & \cmark & \xmark & 66.5 \\
w/o Adapter + SAM$^{*}$           & \xmark & \cmark & 65.4 \\
\textbf{Adapter + SAM$^{*}$}               & \cmark & \cmark & \textbf{76.2} \\
\bottomrule
\end{tabular}
}
\vspace{-4mm}
\end{table}
% \vspace{-10pt}

\textbf{Candidate selector design.}
We analyze the design choices of the proposed Candidate Selector on the RoboTools dataset, as summarized in Table~\ref{tab:selector_ablation}. Here, we do not consider instance-specific object tokens and use the basic SAM model for mask generation, in order to isolate the effect of the Candidate Selector. When the Candidate Selector is removed, the \emph{w/o Filtering} setting directly uses all candidate points produced by dense feature matching as prompts for SAM to generate masks.
\emph{Filtering}, where candidate points are selected based on their dense matching scores following Eq.~\eqref{eq:threshold_filtering}, leads to a certain performance improvement compared to using all candidate points. This indicates that filtering low-confidence matches can partially suppress noisy prompts.
Despite this improvement, the performance of score-based filtering remains significantly inferior to that achieved with the full Candidate Selector, demonstrating the importance of the Candidate Selector, that suppresses false positives caused by local appearance ambiguities beyond simple score thresholding.

We further evaluate the complete Candidate Selector by comparing different visual encoders used for candidate embedding extraction.
The \emph{DINO-CLS} encoder uses the class token generated by DINOv3~\cite{simeoni2025dinov3} as the candidate embedding. In contrast, the \emph{Perception Encoder (PE)}~\cite{bolya2025PerceptionEncoder} consistently achieves better performance.
This suggests that PE provides stronger semantic representations, enabling better association between local object parts and the overall instance appearance, which leads to improved novel instance detection performance.

% \begin{table}[t]
% \centering
% \caption{Ablation study of the Candidate Selector design.}
% \label{tab:selector_ablation}
% \scalebox{1.0}{
% \begin{tabular}{lccc}
% \toprule
% \textbf{Configuration} & \textbf{AP} & \textbf{AP$_{50}$} & \textbf{AP$_{75}$} \\
% \midrule
% \multicolumn{4}{c}{\textit{Without Candidate Selector}} \\
% \midrule
% w/o Score Filtering        & 48.7 & 60.5 & 51.6 \\
% With Score Filtering       & 53.4 & 64.6 & 58.1 \\
% \midrule
% \multicolumn{4}{c}{\textit{With Candidate Selector}} \\
% \midrule
% DINO-CLS Encoder~\cite{simeoni2025dinov3}           & 62.7 & 75.4 & 68.0 \\
% Perception Encoder (PE)~\cite{bolya2025PerceptionEncoder}    & \textbf{64.8} & \textbf{77.6} & \textbf{70.2} \\
% \bottomrule
% \end{tabular}
% }
% \vspace{-2mm}
% \end{table}

\begin{table}[!htbp]
\centering
\caption{Ablation study of the Candidate Selector design.}
\label{tab:selector_ablation}

\makebox[\linewidth][c]{%
\resizebox{1\linewidth}{!}{%
\begin{tabular}{@{}c@{\hspace{10pt}}c@{}} % ← 中间安全间距：18pt（可调）
% -------- left mini table --------
\begin{minipage}[t]{0.68\linewidth}
\newsavebox{\tblA}
\sbox{\tblA}{%
\begin{tabular}{lccc}
\toprule
\textbf{Configuration} & \textbf{AP} & \textbf{AP$_{50}$} & \textbf{AP$_{75}$} \\
\midrule
w/o Filtering~\cite{sun2021loftr}  & 48.7 & 60.5 & 51.6 \\
With Filtering~\cite{oquab2023dinov2}  & 53.4 & 64.6 & 58.1 \\
\bottomrule
\end{tabular}
}
\centering
\makebox[\wd\tblA][c]{\small \textbf{(1) w/o Candidate Selector }}\\[3pt] % ← 调大：3pt/4pt；调小：1pt
\usebox{\tblA}
\end{minipage}
&
% -------- right mini table --------
\begin{minipage}[t]{0.68\linewidth}
\newsavebox{\tblB}
\sbox{\tblB}{%
\begin{tabular}{lccc}
\toprule
\textbf{Encoder} & \textbf{AP} & \textbf{AP$_{50}$} & \textbf{AP$_{75}$} \\
\midrule
DINO-CLS~\cite{simeoni2025dinov3}  & 62.7 & 75.4 & 68.0 \\
\textbf{PE~\cite{bolya2025PerceptionEncoder}}   & \textbf{64.8} & \textbf{77.6} & \textbf{70.2} \\
\bottomrule
\end{tabular}
}
\centering
\makebox[\wd\tblB][c]{\small \textbf{(2) With Candidate Selector } }\\[3pt]
\usebox{\tblB}
\end{minipage}
\end{tabular}
}%
}
% \vspace{-5mm}
\end{table}

\begin{figure}[t]
  \centering
  \begin{minipage}[t]{0.49\linewidth}
    \centering
    \vspace{0pt} % ⭐ 强制顶部对齐
    \includegraphics[width=\linewidth]{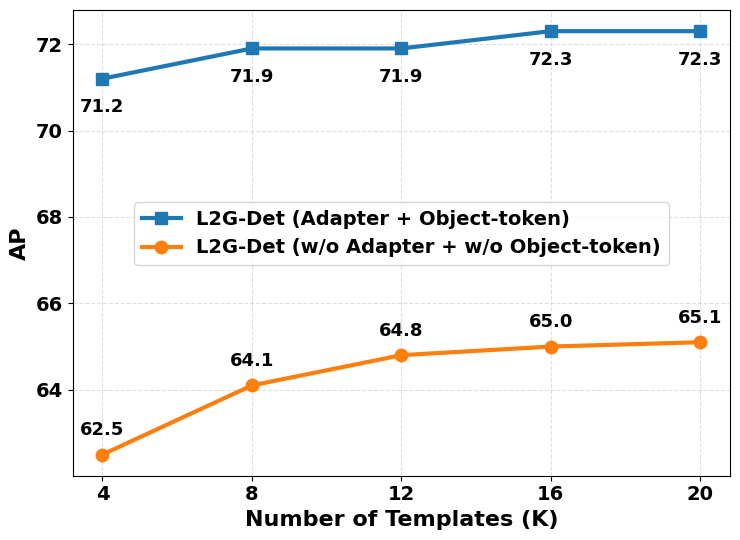}
    \captionof{figure}{\small
    Effect of the number of template images $K$.}
    \label{fig:templates_and_training}
  \end{minipage}
  \hfill
  \begin{minipage}[t]{0.49\linewidth}
  
    \centering
    \vspace{2pt} % ⭐ 强制顶部对齐

    \captionof{table}{\small Comparison of different training strategies.}
    \label{tab:ablation_training_strategy}
    
    \resizebox{1.0\linewidth}{!}{
    \renewcommand{\arraystretch}{1.7}
    \begin{tabular}{lccc}
      \toprule
      \textbf{Training Strategy} & \textbf{AP} & \textbf{AP$_{50}$} & \textbf{AP$_{75}$} \\
      \midrule
      CL-Joint  & 69.5 & 82.7 & 74.8 \\
   
      Joint      & 69.9 & 82.6 & 75.3 \\
   
      \textbf{SAM$^{*}$ (Ours)} & \textbf{71.9} & \textbf{84.6} & \textbf{77.2} \\
      \bottomrule
    \end{tabular}}
    \vspace{2pt} % 
  \end{minipage}

  % \caption{Effect of template quantity and training strategy on detection performance.}
  \vspace{-4mm}
\end{figure}

\textbf{Effect of the number of template images.}
We study the impact of the number of template images $K$ on detection performance on the RoboTools dataset.
As shown in Fig.~\ref{fig:templates_and_training}, increasing $K$ consistently improves performance for both the basic L2G-Det framework and the full model.
This trend indicates that incorporating more template observations provides richer appearance coverage of the target instance, leading to more reliable local matching and candidate generation.

For the baseline L2G-Det framework, we observe that performance gains become marginal once the number of template images exceeds $K=12$.
The full L2G-Det model equipped with the adapter and instance-specific object tokens reaches a high performance level at $K=8$, after which the improvement gradually saturates.
Overall, considering the trade-off between detection accuracy and the cost of collecting template images, we adopt $K=12$ as the default number of templates for L2G-Det. This choice achieves near-peak AP while significantly reducing the reliance on additional template observations.

% \textbf{Training strategies.}
% We compare different training paradigms on RoboTools in Table~\ref{tab:ablation_training_strategy}.
% \emph{Joint} represents the conventional strategy that mixes training data from all 20 objects and optimizes a single model jointly.
% \emph{CL-Joint} evaluates a continual-learning setting~\cite{li2017learning} by splitting the 20 objects into four sequential tasks (objects~1--5, 6--10, 11--15, and 16--20). Training proceeds task-by-task in order, while objects within the same task are still jointly trained.
% Finally, SAM$^{*}$(\emph{Augmented SAM}) corresponds to our proposed instance-specific object-token training, where each object is associated with its own learnable token.
% As shown in Table~\ref{tab:ablation_training_strategy}, SAM$^{*}$ achieves the best performance, indicating that instance-specific tokenization provides a more effective and scalable training strategy than joint-training baselines.

\textbf{Training strategies.}
We compare different training paradigms on RoboTools in Table~\ref{tab:ablation_training_strategy}.
\emph{Joint} represents the conventional strategy that mixes training data from all 20 objects and optimizes a single model jointly.
\emph{CL-Joint} evaluates a continual-learning setting~\cite{li2017learning} by splitting the 20 objects into four sequential tasks (objects~1--5, 6--10, 11--15, and 16--20). 
Training proceeds task-by-task in order, while objects within the same task are still jointly trained.
In an additional per-task analysis, earlier objects show a slight performance drop after learning subsequent tasks; for example, the AP of objects~1--5 decreases from 70.5 to 69.9.
Finally, SAM$^{*}$ (\emph{Augmented SAM}) corresponds to our proposed instance-specific object-token training, where each object is associated with its own learnable token.
As shown in Table~\ref{tab:ablation_training_strategy}, SAM$^{*}$ achieves the best performance, indicating that instance-specific tokenization provides a more effective and scalable training strategy than joint-training baselines.

\textbf{Effect of dense feature extractors.}
We evaluate the impact of different dense feature extractors used in the dense matching stage on the RoboTools dataset, summarized in Table~\ref{tab:dense_backbone_all}.
We compare LoFTR~\cite{sun2021loftr} with large-scale foundation models, including DINOv2-Large~\cite{oquab2023dinov2} and DINOv3-Large~\cite{simeoni2025dinov3}. For the base L2G-Det framework, replacing LoFTR with DINO-based dense features leads to substantial performance improvement, demonstrating the importance of strong semantic representations for reliable local matching. Particularly, DINOv3-Large slightly outperforms DINOv2-Large. These results further indicate that the final instance detection performance is positively correlated with the quality of the dense feature extractor, and validate our choice of DINOv3-Large as the default dense feature extractor in L2G-Det.

\begin{table}[!htbp]
\centering
\caption{Effect of different dense feature extractors used in the dense matching stage.}
\label{tab:dense_backbone_all}

\makebox[\linewidth][c]{%
\resizebox{1\linewidth}{!}{%
\begin{tabular}{@{}c@{\hspace{10pt}}c@{}} % ← 中间安全间距：18pt（可调）
% -------- left mini table --------
\begin{minipage}[t]{0.68\linewidth}
\newsavebox{\tblC}
\sbox{\tblA}{%
\begin{tabular}{lccc}
\toprule
\textbf{Dense Backbone} & \textbf{AP} & \textbf{AP$_{50}$} & \textbf{AP$_{75}$} \\
\midrule
LoFTR~\cite{sun2021loftr}  & 41.3 & 49.2 & 44.4 \\
DINOv2-Large~\cite{oquab2023dinov2}  & 64.4 & 76.3 & 69.8 \\
\textbf{DINOv3-Large~\cite{simeoni2025dinov3}} & \textbf{64.8} & \textbf{77.6} & \textbf{70.2} \\
\bottomrule
\end{tabular}
}
\centering
\makebox[\wd\tblA][c]{\small \textbf{(1) L2G-Det w/o Adapter + SAM }}\\[3pt] % ← 调大：3pt/4pt；调小：1pt
\usebox{\tblA}
\end{minipage}
&
% -------- right mini table --------
\begin{minipage}[t]{0.68\linewidth}
\newsavebox{\tblD}
\sbox{\tblB}{%
\begin{tabular}{lccc}
\toprule
\textbf{Dense Backbone} & \textbf{AP} & \textbf{AP$_{50}$} & \textbf{AP$_{75}$} \\
\midrule
LoFTR~\cite{sun2021loftr}  & 48.3 & 56.8 & 51.7 \\
DINOv2-Large~\cite{oquab2023dinov2}  & 71.4 & 83.8 & 76.6 \\
\textbf{DINOv3-Large~\cite{simeoni2025dinov3}} & \textbf{71.9} & \textbf{84.6} & \textbf{77.2} \\
\bottomrule
\end{tabular}
}
\centering
\makebox[\wd\tblB][c]{\small \textbf{(2) L2G-Det with Adapter + SAM$^{*}$ } }\\[3pt]
\usebox{\tblB}
\end{minipage}
\end{tabular}
}%
}
% \vspace{-1mm}
\end{table}

\textbf{Effect of backbone choice.}
To ensure a fair comparison with other instance detection methods, we use the same SAM ViT-Large backbone as the compared baselines and vary the DINO backbone. Specifically, we compare L2G-Det with the same or smaller DINOv2~\cite{oquab2023dinov2} backbones.
As shown in Table~\ref{tab:backbone_fairness}, L2G-Det consistently outperforms the compared methods even when using the same or smaller DINOv2 backbones.
In particular, L2G-Det with DINOv2-Small already achieves higher performance than NIDS-Net~\cite{lu2024adapting} with DINOv2-Large.
This indicates that, although stronger dense features can further improve performance, the main improvement comes from the proposed local-to-global pipeline design rather than simply using a larger pretrained backbone.

\begin{table}[!htbp]
\centering
\caption{Effect of backbone choice on RoboTools benchmark with the adapter and augmented SAM.}
\label{tab:backbone_fairness}
\resizebox{0.85\linewidth}{!}{
\begin{tabular}{lcccc}
\toprule
\textbf{Method} & \textbf{Backbone} & \textbf{AP} & \textbf{AP$_{50}$} & \textbf{AP$_{75}$} \\
\midrule
IDOW~\cite{shen2025solving} & DINOv2-Small & 51.9 & 63.8 & 56.5 \\
NIDS-Net~\cite{lu2024adapting} & DINOv2-Large & 64.9 & 79.4 & 70.8 \\
L2G-Det (Ours) & DINOv2-Small & 69.2 & 81.1 & 74.3 \\
\textbf{L2G-Det (Ours)} & \textbf{DINOv2-Base} & \textbf{70.8} & \textbf{82.9} & \textbf{76.1} \\
\bottomrule
\end{tabular}
}
\vspace{-3mm}
\end{table}

\subsection{Real-World Robotic Experiments}

\begin{figure}
    \centering
    \includegraphics[width=\linewidth]{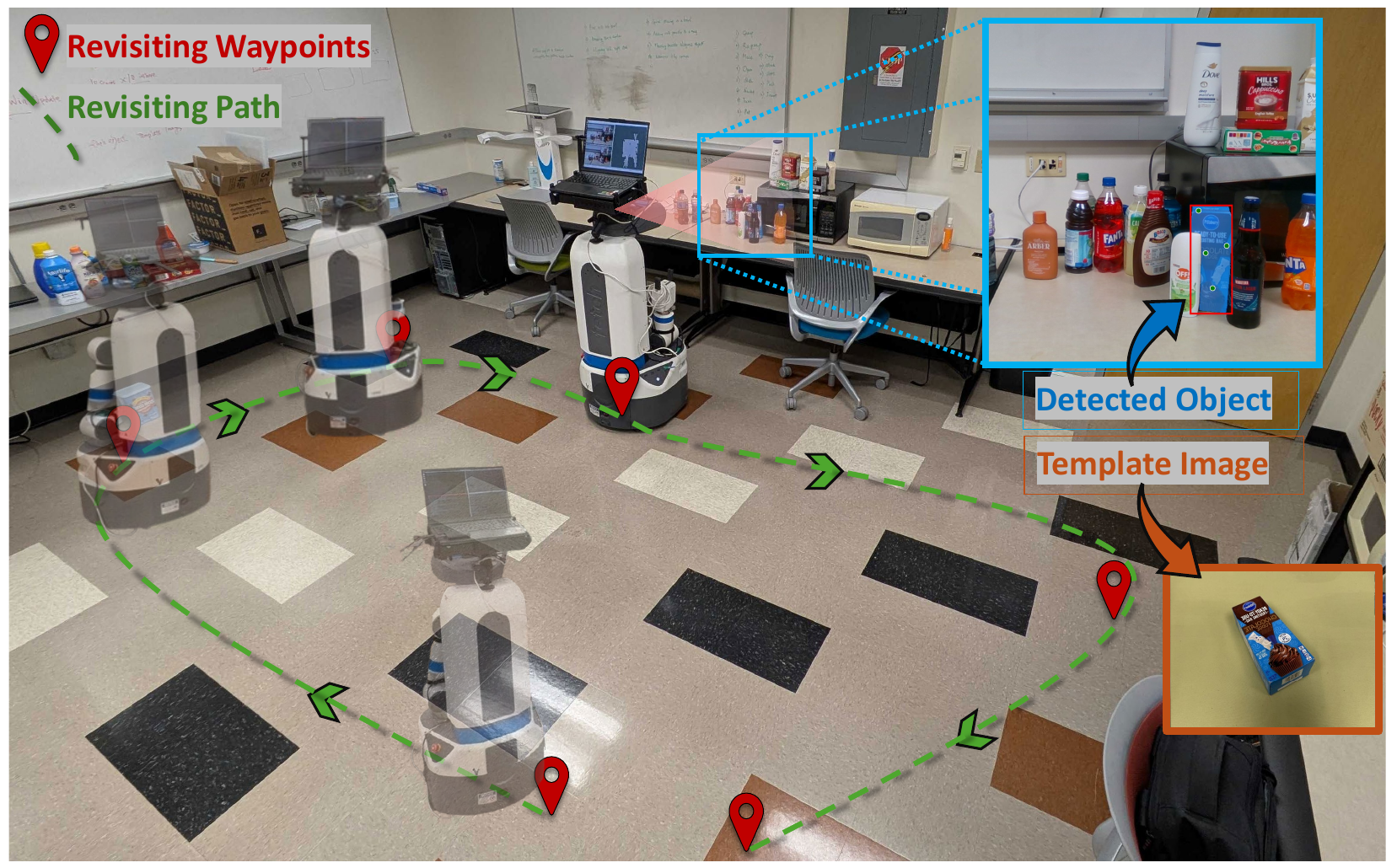}
    \caption{Overview of the robot experiment process showing the waypoint tracking and target object identification.}
    \label{fig:robot-experiment}
    \vspace{-4mm}
\end{figure}

% To evaluate the practical applicability of our approach in real-world open-world settings, we deploy our system on a mobile robot equipped with an RGB camera and a navigation stack. Given a set of template images for a novel target object, the robot is tasked with searching for and localizing the object in cluttered indoor environments.

To evaluate the applicability of our approach in open-world settings, we deploy our system on an RTX 4090 laptop, mounted on a Fetch robot, connected via Ethernet. Given the template images of  novel target objects, the robot searches for and localizes the objects in cluttered indoor environment. For inference, we mount a Realsense D415 camera on the robot head and stream the RGB images (1280$\times$720, 15fps).

We first map the environment with ROS Gmapping~\cite{rosgmapping}, record the robot trajectory and extract a revisit path of 6 waypoints using the procedure mentioned in ~\cite{allu2025modularroboticautonomousexploration}. Next, the robot localizes using AMCL ROS~\cite{navigation_stack} and revisits these waypoints, while we run real-time RGB inference (Fig.~\ref{fig:robot-experiment}). We tested 8 objects with 2 settings: 1) Augmented SAM with instance-specific object tokens; 2) using the pretrained SAM. Based on the predicted mask, we compute the matching score using cosine similarity following Eq.~\eqref{eq:matching_score}.
If the matching score exceeds the threshold (here 0.7), the corresponding detection is accepted, and the trial stops at that waypoint. Table~\ref{tab:robot_results} shows that \textbf{L2G-Det} robustly detects all 8 objects in real-world environments. The augmented SAM yields more accurate and complete masks, improving performance under stricter IoU thresholds. Experimental videos are included on our project website: \url{https://irvlutd.github.io/L2G/}.

\begin{table}[t]
\centering
\caption{Real-world robotic detection results on 8 objects. $N_{\text{IoU}>\tau}$ denotes the number of detections whose predicted masks achieve IoU $>\tau$ (threshold)  w.r.t ground-truth annotations. $N_{\text{Success}}$ denotes the number of trials in which the robot successfully
detects the target object and stops. There are 8 trials in total.}
\label{tab:robot_results}
\scalebox{0.85}{
\begin{tabular}{lcccc}
\toprule
\textbf{Method} 
& $\mathbf{N_{\mathrm{IoU}>0.5}}$
& $\mathbf{N_{\mathrm{IoU}>0.75}}$
& $\mathbf{N_{\mathrm{IoU}>0.95}}$
& $\mathbf{N_{\mathrm{Success}}}$ \\
\midrule
L2G-Det with SAM            & 7 & 7 & 7 & 8 \\
\textbf{L2G-Det with SAM$^{*}$}  & \textbf{8} & \textbf{8} & \textbf{7} & \textbf{8} \\
\bottomrule
\end{tabular}
}
\vspace{-4mm}
\end{table}

% \begin{figure}

%   \centering
%   \includegraphics[width=1.1\linewidth]{realworld.pdf}
%   \caption{Instance detection results on real-world benchmark.}
% \vspace{-1mm}
%   \label{fig}
% \end{figure}

\subsection{Computational Cost}

We further report the computational cost of L2G-Det in Table~\ref{tab:cost_analysis}. 
The real-world robotic experiments are deployed on a laptop with an NVIDIA RTX 4090 GPU, while the benchmark experiments on RoboTools and HR-InsDet are conducted using two NVIDIA RTX A5000 GPUs for training and inference. 
The reported inference runtime is the average end-to-end per-object cost measured over the entire benchmark. 
Since we sample only $S=10$ patches per template image and use a small number of templates $K$, the number of dense matching features remains limited. 
While the current implementation still has non-trivial computational overhead, the improved detection accuracy suggests that the proposed local-to-global formulation provides a practical and effective solution for open-world robotic perception.

% In the Candidate Selector, SAM is invoked for multiple candidate points, but its image encoder is executed only once per query image, keeping the overall inference cost manageable.

\begin{table}[!htbp]
\centering
\caption{Summary of training-related and inference cost. The inference runtime is reported as the end-to-end per-object cost with $K=4$ template images.}
\label{tab:cost_analysis}
\scalebox{0.95}{
\begin{tabular}{lcc}
\toprule
\textbf{Training-related Cost} & \textbf{Runtime} & \textbf{GPU Memory} \\
\midrule
Create synthetic images (per object) & 3 min & $\varnothing$ \\
Train object token (per object) & 25 min & 2400 MiB \\
Train Adapter (all objects) & 2 h & 6900 MiB \\
\midrule
\textbf{Inference Cost ($K=4$)} & \textbf{Runtime} & \textbf{GPU Memory} \\
\midrule
RoboTools (1920$\times$1080) & 1.10 s & 8300 MiB \\
HR-InsDet (8192$\times$6144) & 4.10 s & 20000 MiB \\
\bottomrule
\end{tabular}
}
\vspace{-3mm}
\end{table}

Table~\ref{tab:k_runtime} further reports the inference cost under different numbers of template images $K$ on RoboTools. 
This reflects the trade-off between efficiency and performance: using fewer templates reduces inference time, while using more templates can improve robustness in the zero-shot setting at the cost of additional runtime.

\begin{table}[!htbp]
\centering
\caption{Inference runtime and GPU memory with different numbers of template images $K$ on RoboTools.}
\label{tab:k_runtime}
\scalebox{0.95}{
\begin{tabular}{lcc}
\toprule
\textbf{$K$} & \textbf{Runtime} & \textbf{GPU Memory} \\
\midrule
4  & 1.1 s & 8300 MiB \\
8  & 1.6 s & 9000 MiB \\
12 & 1.9 s & 9800 MiB \\
16 & 2.1 s & 10500 MiB \\
\bottomrule
\end{tabular}
}
\vspace{-3mm}
\end{table}

\section{Conclusion} 
\label{sec:conclusion}

We introduced L2G-Det, a local-to-global framework for novel object instance detection and segmentation in open-world environments. By replacing proposal-based pipelines with dense patch-level matching and mask reconstruction, our approach robustly detects novel objects under occlusion, viewpoint variation, and background clutter. We further proposed a candidate selection module to suppress false positives and an augmented SAM with instance-specific object tokens to recover coherent global masks from sparse prompts. Experimental results on benchmark datasets and real-world robotic scenarios demonstrate the effectiveness and robustness of the proposed approach, highlighting its potential for scalable instance-level perception in open-world robotics.

\textbf{Limitations.} First, our framework integrates multiple pre-trained models for dense feature extraction, candidate selection, and mask generation, rather than employing a fully end-to-end detector. As a result, it requires higher computational resources compared to conventional end-to-end detection pipelines.
Second, the learning of instance-specific object tokens relies on template-based synthetic training images, where target objects are generated by simple copy-and-paste operations onto background images, which may not fully capture complex real-world interactions. Exploring generative models to produce more realistic training images is an important direction for future work.

% \textbf{Limitations.} \yu{Write some limitations}

\section*{Acknowledgments}

This work was supported in part by the National Science Foundation (NSF) under Grant Nos. 2346528 and 2520553, and the NVIDIA Academic Grant Program Award, and a gift funding from XPeng.

%% Use plainnat to work nicely with natbib. 

\bibliographystyle{plainnat}
\bibliography{references}

@article{ammirato2018targetInstDet,
  title={Target driven instance detection},
  author={Ammirato, Phil and Fu, Cheng-Yang and Shvets, Mykhailo and Kosecka, Jana and Berg, Alexander C},
  journal={arXiv preprint arXiv:1803.04610},
  year={2018}
}

@inproceedings{mercier2021deepTemplate,
  title={Deep template-based object instance detection},
  author={Mercier, Jean-Philippe and Garon, Mathieu and Giguere, Philippe and Lalonde, Jean-Francois},
  booktitle={Proceedings of the IEEE/CVF Winter Conference on Applications of Computer Vision},
  pages={1507--1516},
  year={2021}
}

@article{li2024voxdet,
  title={VoxDet: Voxel Learning for Novel Instance Detection},
  author={Li, Bowen and Wang, Jiashun and Hu, Yaoyu and Wang, Chen and Scherer, Sebastian},
  journal={Advances in Neural Information Processing Systems},
  volume={36},
  year={2024}
}

@inproceedings{liu2022gen6d,
  title={Gen6d: Generalizable model-free 6-dof object pose estimation from rgb images},
  author={Liu, Yuan and Wen, Yilin and Peng, Sida and Lin, Cheng and Long, Xiaoxiao and Komura, Taku and Wang, Wenping},
  booktitle={European Conference on Computer Vision},
  pages={298--315},
  year={2022},
  organization={Springer}
}

@inproceedings{caron2021dino,
  title={Emerging Properties in Self-Supervised Vision Transformers},
  author={Caron, Mathilde and Touvron, Hugo and Misra, Ishan and J\'egou, Herv\'e  and Mairal, Julien and Bojanowski, Piotr and Joulin, Armand},
  booktitle={Proceedings of the International Conference on Computer Vision (ICCV)},
  year={2021}
}

@article{oquab2023dinov2,
  title={Dinov2: Learning robust visual features without supervision},
  author={Oquab, Maxime and Darcet, Timoth{\'e}e and Moutakanni, Th{\'e}o and Vo, Huy and Szafraniec, Marc and Khalidov, Vasil and Fernandez, Pierre and Haziza, Daniel and Massa, Francisco and El-Nouby, Alaaeldin and others},
  journal={arXiv:2304.07193},
  year={2023}
}

@inproceedings{radford2021clip,
  title={Learning transferable visual models from natural language supervision},
  author={Radford, Alec and Kim, Jong Wook and Hallacy, Chris and Ramesh, Aditya and Goh, Gabriel and Agarwal, Sandhini and Sastry, Girish and Askell, Amanda and Mishkin, Pamela and Clark, Jack and others},
  booktitle={International conference on machine learning},
  pages={8748--8763},
  year={2021},
  organization={PMLR}
}

@article{gao2021clipAdapter,
  title={Clip-adapter: Better vision-language models with feature adapters},
  author={Gao, Peng and Geng, Shijie and Zhang, Renrui and Ma, Teli and Fang, Rongyao and Zhang, Yongfeng and Li, Hongsheng and Qiao, Yu},
  journal={arXiv 2110.04544},
  year={2021}
}

@inproceedings{kirillov2023SAM,
  title={Segment anything},
  author={Kirillov, Alexander and Mintun, Eric and Ravi, Nikhila and Mao, Hanzi and Rolland, Chloe and Gustafson, Laura and Xiao, Tete and Whitehead, Spencer and Berg, Alexander C and Lo, Wan-Yen and others},
  booktitle={Proceedings of the IEEE/CVF International Conference on Computer Vision},
  pages={4015--4026},
  year={2023}
}

@article{dosovitskiy2020ViT,
  title={An image is worth 16x16 words: Transformers for image recognition at scale},
  author={Dosovitskiy, Alexey and Beyer, Lucas and Kolesnikov, Alexander and Weissenborn, Dirk and Zhai, Xiaohua and Unterthiner, Thomas and Dehghani, Mostafa and Minderer, Matthias and Heigold, Georg and Gelly, Sylvain and others},
  journal={arXiv:2010.11929},
  year={2020}
}

@article{liu2023groundingDINO,
  title={Grounding dino: Marrying dino with grounded pre-training for open-set object detection},
  author={Liu, Shilong and Zeng, Zhaoyang and Ren, Tianhe and Li, Feng and Zhang, Hao and Yang, Jie and Li, Chunyuan and Yang, Jianwei and Su, Hang and Zhu, Jun and others},
  journal={arXiv preprint arXiv:2303.05499},
  year={2023}
}

@inproceedings{shen2023instance,
  title={A High-Resolution Dataset for Instance Detection with Multi-View Instance Capture},
  author={Shen, Qianqian and Zhao, Yunhan and Kwon, Nahyun and Kim, Jeeeun and Li, Yanan and Kong, Shu},
  booktitle={NeurIPS Datasets and Benchmarks Track},
  year={2023}
}

@article{ren2015faster,
  title={Faster r-cnn: Towards real-time object detection with region proposal networks},
  author={Ren, Shaoqing and He, Kaiming and Girshick, Ross and Sun, Jian},
  journal={Advances in neural information processing systems},
  volume={28},
  year={2015}
}

@inproceedings{lin2017focal,
  title={Focal loss for dense object detection},
  author={Lin, Tsung-Yi and Goyal, Priya and Girshick, Ross and He, Kaiming and Doll{\'a}r, Piotr},
  booktitle={Proceedings of the IEEE international conference on computer vision},
  pages={2980--2988},
  year={2017}
}

@article{zhou2019objects,
  title={Objects as points},
  author={Zhou, Xingyi and Wang, Dequan and Kr{\"a}henb{\"u}hl, Philipp},
  journal={arXiv preprint arXiv:1904.07850},
  year={2019}
}

@INPROCEEDINGS{9010746,
  author={Tian, Zhi and Shen, Chunhua and Chen, Hao and He, Tong},
  booktitle={2019 IEEE/CVF International Conference on Computer Vision (ICCV)}, 
  title={FCOS: Fully Convolutional One-Stage Object Detection}, 
  year={2019},
  volume={},
  number={},
  pages={9626-9635},
  doi={10.1109/ICCV.2019.00972}
}

@misc{zhang2022dinodetr,
      title={DINO: DETR with Improved DeNoising Anchor Boxes for End-to-End Object Detection}, 
      author={Hao Zhang and Feng Li and Shilong Liu and Lei Zhang and Hang Su and Jun Zhu and Lionel M. Ni and Heung-Yeung Shum},
      year={2022},
      eprint={2203.03605},
      archivePrefix={arXiv},
      primaryClass={cs.CV}
}

@article{kim2022learning,
  title={Learning open-world object proposals without learning to classify},
  author={Kim, Dahun and Lin, Tsung-Yi and Angelova, Anelia and Kweon, In So and Kuo, Weicheng},
  journal={IEEE Robotics and Automation Letters},
  volume={7},
  number={2},
  pages={5453--5460},
  year={2022},
  publisher={IEEE}
}

@inproceedings{osokin2020os2d,
  title={Os2d: One-stage one-shot object detection by matching anchor features},
  author={Osokin, Anton and Sumin, Denis and Lomakin, Vasily},
  booktitle={Computer Vision--ECCV 2020: 16th European Conference, Glasgow, UK, August 23--28, 2020, Proceedings, Part XV 16},
  pages={635--652},
  year={2020},
  organization={Springer}
}

@inproceedings{nguyen2023cnos,
title={CNOS: A Strong Baseline for CAD-based Novel Object Segmentation},
author={Nguyen, Van Nguyen and Groueix, Thibault and Ponimatkin, Georgy and Lepetit, Vincent and Hodan, Tomas},
booktitle={Proceedings of the IEEE/CVF International Conference on Computer Vision},
pages={2134--2140},
year={2023}
}

@article{oord2018representation,
  title={Representation learning with contrastive predictive coding},
  author={Oord, Aaron van den and Li, Yazhe and Vinyals, Oriol},
  journal={arXiv preprint arXiv:1807.03748},
  year={2018}
}

@inproceedings{chen2020simple,
  title={A simple framework for contrastive learning of visual representations},
  author={Chen, Ting and Kornblith, Simon and Norouzi, Mohammad and Hinton, Geoffrey},
  booktitle={International conference on machine learning},
  pages={1597--1607},
  year={2020},
  organization={PMLR}
}

@article{kingma2014adam,
  title={Adam: A method for stochastic optimization},
  author={Kingma, Diederik P and Ba, Jimmy},
  journal={arXiv preprint arXiv:1412.6980},
  year={2014}
}

@misc{rosgmapping,
  author       = { Brian Gerkey},
  title        = {slam\_gmapping},
  year         = {2013},
  URL = {https://github.com/ros-perception/slam\_gmapping},
}

@misc{allu2025modularroboticautonomousexploration,
      title={A Modular Robotic System for Autonomous Exploration and Semantic Updating in Large-Scale Indoor Environments}, 
      author={Sai Haneesh Allu and Itay Kadosh and Tyler Summers and Yu Xiang},
      year={2025},
      eprint={2409.15493},
      archivePrefix={arXiv},
      primaryClass={cs.RO},
      url={https://arxiv.org/abs/2409.15493}, 
}

@misc{navigation_stack,
  title = {ROS Navigation Stack},
  author = {{Eitan Marder-Eppstein} and {David V. Lu} and {Michael Ferguson} and {Aaron Hoy}},
  URL = {https://github.com/ros-planning/navigation},
}

@misc{simeoni2025dinov3,
  title={{DINOv3}},
  author={Sim{\'e}oni, Oriane and Vo, Huy V. and Seitzer, Maximilian and Baldassarre, Federico and Oquab, Maxime and Jose, Cijo and Khalidov, Vasil and Szafraniec, Marc and Yi, Seungeun and Ramamonjisoa, Micha{\"e}l and Massa, Francisco and Haziza, Daniel and Wehrstedt, Luca and Wang, Jianyuan and Darcet, Timoth{\'e}e and Moutakanni, Th{\'e}o and Sentana, Leonel and Roberts, Claire and Vedaldi, Andrea and Tolan, Jamie and Brandt, John and Couprie, Camille and Mairal, Julien and J{\'e}gou, Herv{\'e} and Labatut, Patrick and Bojanowski, Piotr},
  year={2025},
  eprint={2508.10104},
  archivePrefix={arXiv},
  primaryClass={cs.CV},
  url={https://arxiv.org/abs/2508.10104},
}

@article{ravi2024sam2,
  title={SAM 2: Segment Anything in Images and Videos},
  author={Ravi, Nikhila and Gabeur, Valentin and Hu, Yuan-Ting and Hu, Ronghang and Ryali, Chaitanya and Ma, Tengyu and Khedr, Haitham and R{\"a}dle, Roman and Rolland, Chloe and Gustafson, Laura and Mintun, Eric and Pan, Junting and Alwala, Kalyan Vasudev and Carion, Nicolas and Wu, Chao-Yuan and Girshick, Ross and Doll{\'a}r, Piotr and Feichtenhofer, Christoph},
  journal={arXiv preprint arXiv:2408.00714},
  url={https://arxiv.org/abs/2408.00714},
  year={2024}
}

@article{bolya2025PerceptionEncoder,
  title={Perception Encoder: The best visual embeddings are not at the output of the network},
  author={Daniel Bolya and Po-Yao Huang and Peize Sun and Jang Hyun Cho and Andrea Madotto and Chen Wei and Tengyu Ma and Jiale Zhi and Jathushan Rajasegaran and Hanoona Rasheed and Junke Wang and Marco Monteiro and Hu Xu and Shiyu Dong and Nikhila Ravi and Daniel Li and Piotr Doll{\'a}r and Christoph Feichtenhofer},
  journal={arXiv:2504.13181},
  year={2025}
}

@misc{lu2024adapting,
      title={Adapting Pre-Trained Vision Models for Novel Instance Detection and Segmentation}, 
      author={Yangxiao Lu and Jishnu Jaykumar P and Yunhui Guo and Nicholas Ruozzi and Yu Xiang},
      year={2024},
      eprint={2405.17859},
      archivePrefix={arXiv},
      primaryClass={cs.CV}
}

@inproceedings{shen2025solving,
  title={Solving Instance Detection from an Open-World Perspective},
  author={Shen, Qianqian and Zhao, Yunhan and Kwon, Nahyun and Kim, Jeeeun and Li, Yanan and Kong, Shu},
  booktitle={Proceedings of the IEEE/CVF Conference on Computer Vision and Pattern Recognition (CVPR)},
  year={2025}
}

@inproceedings{lester2021power,
  title     = {The Power of Scale for Parameter-Efficient Prompt Tuning},
  author    = {Lester, Brian and Al-Rfou, Rami and Constant, Noah},
  booktitle = {Proceedings of the 2021 Conference on Empirical Methods in Natural Language Processing (EMNLP)},
  year      = {2021},
  pages     = {3045--3059}
}

@inproceedings{jia2022visual,
  title     = {Visual Prompt Tuning},
  author    = {Jia, Menglin and Tang, Luming and Chen, Bor-Chun and Cardie, Claire and Belongie, Serge},
  booktitle = {Proceedings of the European Conference on Computer Vision (ECCV)},
  year      = {2022},
  pages     = {709--727}
}

@article{tian2020makes,
  title={What makes for good views for contrastive learning?},
  author={Tian, Yonglong and Sun, Chen and Poole, Ben and Krishnan, Dilip and Schmid, Cordelia and Isola, Phillip},
  journal={Advances in neural information processing systems},
  volume={33},
  pages={6827--6839},
  year={2020}
}

@article{sun2021loftr,
  title={{LoFTR}: Detector-Free Local Feature Matching with Transformers},
  author={Sun, Jiaming and Shen, Zehong and Wang, Yuang and Bao, Hujun and Zhou, Xiaowei},
  journal={CVPR},
  year={2021}
}

@book{goodfellow2016deep,
  title={Deep Learning},
  author={Goodfellow, Ian and Bengio, Yoshua and Courville, Aaron},
  publisher={MIT Press},
  year={2016}
}

@inproceedings{milletari2016vnet,
  title={V-Net: Fully Convolutional Neural Networks for Volumetric Medical Image Segmentation},
  author={Milletari, Fausto and Navab, Nassir and Ahmadi, Seyed-Ahmad},
  booktitle={2016 Fourth International Conference on 3D Vision (3DV)},
  year={2016}
}

@inproceedings{rahman2016iou,
  title={Optimizing Intersection-over-Union in Deep Neural Networks for Image Segmentation},
  author={Rahman, Md. Asiful Islam and Wang, Yang},
  booktitle={International Symposium on Visual Computing (ISVC)},
  year={2016}
}

@inproceedings{dwibedi2017cut,
  title={Cut, Paste and Learn: Surprisingly Easy Synthesis for Instance Detection},
  author={Dwibedi, Debidatta and Misra, Ishan and Hebert, Martial},
  booktitle={Proceedings of the IEEE International Conference on Computer Vision (ICCV)},
  year={2017}
}

@article{lowe2004distinctive,
  title={Distinctive Image Features from Scale-Invariant Keypoints},
  author={Lowe, David G.},
  journal={International Journal of Computer Vision (IJCV)},
  year={2004},
  volume={60},
  number={2},
  pages={91--110}
}

@inproceedings{bay2006surf,
  title={SURF: Speeded Up Robust Features},
  author={Bay, Herbert and Tuytelaars, Tinne and Van Gool, Luc},
  booktitle={European Conference on Computer Vision (ECCV)},
  year={2006},
  pages={404--417}
}

@inproceedings{wang2024eloftr,
  title={{Efficient LoFTR}: Semi-Dense Local Feature Matching with Sparse-Like Speed},
  author={Wang, Yifan and He, Xingyi and Peng, Sida and Tan, Dongli and Zhou, Xiaowei},
  booktitle={CVPR},
  year={2024}
}

@misc{carion2025sam3segmentconcepts,
      title={SAM 3: Segment Anything with Concepts},
      author={Nicolas Carion and Laura Gustafson and Yuan-Ting Hu and Shoubhik Debnath and Ronghang Hu and Didac Suris and Chaitanya Ryali and Kalyan Vasudev Alwala and Haitham Khedr and Andrew Huang and Jie Lei and Tengyu Ma and Baishan Guo and Arpit Kalla and Markus Marks and Joseph Greer and Meng Wang and Peize Sun and Roman Rädle and Triantafyllos Afouras and Effrosyni Mavroudi and Katherine Xu and Tsung-Han Wu and Yu Zhou and Liliane Momeni and Rishi Hazra and Shuangrui Ding and Sagar Vaze and Francois Porcher and Feng Li and Siyuan Li and Aishwarya Kamath and Ho Kei Cheng and Piotr Dollár and Nikhila Ravi and Kate Saenko and Pengchuan Zhang and Christoph Feichtenhofer},
      year={2025},
      eprint={2511.16719},
      archivePrefix={arXiv},
      primaryClass={cs.CV},
      url={https://arxiv.org/abs/2511.16719},
}

@inproceedings{li2017learning,
  title={Learning without Forgetting},
  author={Li, Zhizhong and Hoiem, Derek},
  booktitle={IEEE Transactions on Pattern Analysis and Machine Intelligence (TPAMI)},
  year={2017},
  pages={1--1},
  publisher={IEEE}
}

@inproceedings{bormann2021real,
  title={Real-time Instance Detection with Fast Incremental Learning},
  author={Bormann, Richard and Wang, Xinjie and V{\"o}lk, Markus and Kleeberger, Kilian and Lindermayr, Jochen},
  booktitle={Proceedings of the IEEE International Conference on Robotics and Automation (ICRA)},
  year={2021},
  publisher={IEEE},
  doi={10.1109/ICRA48506.2021.9561202}
}

@inproceedings{sarlin2020superglue,
  title={SuperGlue: Learning Feature Matching with Graph Neural Networks},
  author={Sarlin, Paul-Edouard and DeTone, Daniel and Malisiewicz, Tomasz and Rabinovich, Andrew},
  booktitle={Proceedings of the IEEE/CVF Conference on Computer Vision and Pattern Recognition (CVPR)},
  year={2020}
}

@inproceedings{wang2019densefusion,
  title={DenseFusion: 6D Object Pose Estimation by Iterative Dense Fusion},
  author={Wang, Chen and Xu, Danfei and Zhu, Yuke and Mart{\'\i}n-Mart{\'\i}n, Roberto and Lu, Cewu and Fei-Fei, Li and Savarese, Silvio},
  booktitle={Proceedings of the IEEE/CVF Conference on Computer Vision and Pattern Recognition (CVPR)},
  year={2019}
}

@inproceedings{chen2023sam,
  title={Sam-adapter: Adapting segment anything in underperformed scenes},
  author={Chen, Tianrun and Zhu, Lanyun and Deng, Chaotao and Cao, Runlong and Wang, Yan and Zhang, Shangzhan and Li, Zejian and Sun, Lingyun and Zang, Ying and Mao, Papa},
  booktitle={Proceedings of the IEEE/CVF International Conference on Computer Vision},
  pages={3367--3375},
  year={2023}
}

@article{yu2024ts,
  title={TS-SAM: Fine-Tuning Segment-Anything Model for Downstream Tasks},
  author={Yu, Yang and Xu, Chen and Wang, Kai},
  journal={arXiv preprint arXiv:2408.01835},
  year={2024}
}

@inproceedings{kirkpatrick2017overcoming,
  title={Overcoming Catastrophic Forgetting in Neural Networks},
  author={Kirkpatrick, James and Pascanu, Razvan and Rabinowitz, Neil C. and
          Veness, Joel and Desjardins, Guillaume and Rusu, Andrei A. and
          Milan, Kieran and Quan, John and Ramalho, Tiago and Grabska-Barwinska, Agnieszka
          and others},
  booktitle={Proceedings of the National Academy of Sciences (PNAS)},
  year={2017},
  volume={114},
  number={13},
  pages={3521--3526}
}

\clearpage

\appendix
\subsection{More Ablation Studies}

\textbf{Effect of Template-specific Similarity Computation.}
We further conduct an ablation study to analyze the role of template-specific similarity computation in Eq.~\eqref{eq:similar}. 
Our method computes the similarity score between each candidate embedding and its \emph{corresponding} template embedding, which we denote as \textbf{Current\_TEMP (Ours)}.
Specifically, for the $k$-th template image, the candidate embedding $\mathbf{z}_i^k$ and the template embedding $\mathbf{z}_{\text{temp}}^k$ are computed as
\begin{equation}
\mathbf{z}_{i}^{k}
=
\mathcal{A}\!\left(
E\!\left(\mathcal{I}_{\text{query}} \odot \mathcal{M}_{i}^{k}\right)
\right), \tag{3}
\end{equation}
\begin{equation}
\mathbf{z}_{\text{temp}}^{k}
=
\mathcal{A}\!\left(
E\!\left(\mathcal{I}_{\text{temp}}^{k} \odot \mathcal{M}_{\text{temp}}^{k}\right)
\right), \tag{4}
\end{equation}
and their cosine similarity is calculated by

\begin{equation}
\label{eq:similar}
s_i^k
=
\frac{\mathbf{z}_i^k \cdot \mathbf{z}_{\mathrm{temp}}^k}
{\lVert \mathbf{z}_i^k \rVert_2 \, \lVert \mathbf{z}_{\mathrm{temp}}^k \rVert_2}. \tag{7}
\end{equation}
This design preserves the correspondence between each candidate region and the specific template image from which the candidate point is generated.

As a comparison, we introduce an alternative baseline denoted as \textbf{AVG\_TEMP}.
In this setting, instead of using the corresponding template embedding $\mathbf{z}_{\text{temp}}^k$, we first compute the average embedding over all $K$ template images:
\begin{equation}
\mathbf{\bar{z}}_{\text{temp}}
=
\frac{1}{K}
\sum_{k=1}^{K}
\mathcal{A}\!\left(
E\!\left(\mathcal{I}_{\text{temp}}^{k} \odot \mathcal{M}_{\text{temp}}^{k}\right)
\right), \tag{4*}
\end{equation}
and then replace Eq.~\eqref{eq:similar} with
\begin{equation}
\label{eq:similar_avg}
s_i
=
\frac{\mathbf{z}_i^k \cdot \mathbf{\bar{z}}_{\mathrm{temp}}}
{\lVert \mathbf{z}_i^k \rVert_2 \, \lVert \mathbf{\bar{z}}_{\mathrm{temp}} \rVert_2}. \tag{7*}
\end{equation}
All other components and settings remain unchanged.

We analyze the design choices of this component on the RoboTools~\cite{li2024voxdet} dataset. The quantitative results are reported in Table~\ref{tab:ablation_template_avg}. Here, we do not consider instance-specific object tokens and use the basic SAM model~\cite{ravi2024sam2} for mask generation, in order to isolate the effect of this component.
We observe that \textbf{Current\_TEMP (Ours)} consistently outperforms \textbf{AVG\_TEMP}.
This demonstrates that computing similarity with the corresponding template image is more effective than using an averaged template representation.
We attribute this improvement to the fact that different template images capture distinct viewpoints and emphasize different local object parts.
By maintaining template-specific similarity computation, the matching score can more accurately reflect the local appearance consistency between the candidate region and the corresponding template, leading to more reliable candidate selection.
In contrast, averaging template embeddings tends to dilute such distinctive local cues, resulting in reduced discriminative power.

\begin{table}[t]
\centering
\caption{Ablation study on template-specific similarity computation.
\textbf{Current\_TEMP (Ours)} computes similarity using the corresponding template embedding for each candidate,
while \textbf{AVG\_TEMP} uses the averaged embedding over all templates.}
\label{tab:ablation_template_avg}
\begin{tabular}{lccc}
\toprule
\textbf{Method} & \textbf{AP} & \textbf{AP$_{50}$} & \textbf{AP$_{75}$} \\
\midrule
AVG\_TEMP       & 61.4 & 74.2 & 66.6 \\
\textbf{Current\_TEMP (Ours)} & \textbf{64.8} & \textbf{77.6} & \textbf{70.2} \\
\bottomrule
\end{tabular}
\end{table}

\subsection{Qualitative Results}
\label{Appendix:results}
\begin{figure*}

  \centering
  \includegraphics[width=\linewidth]{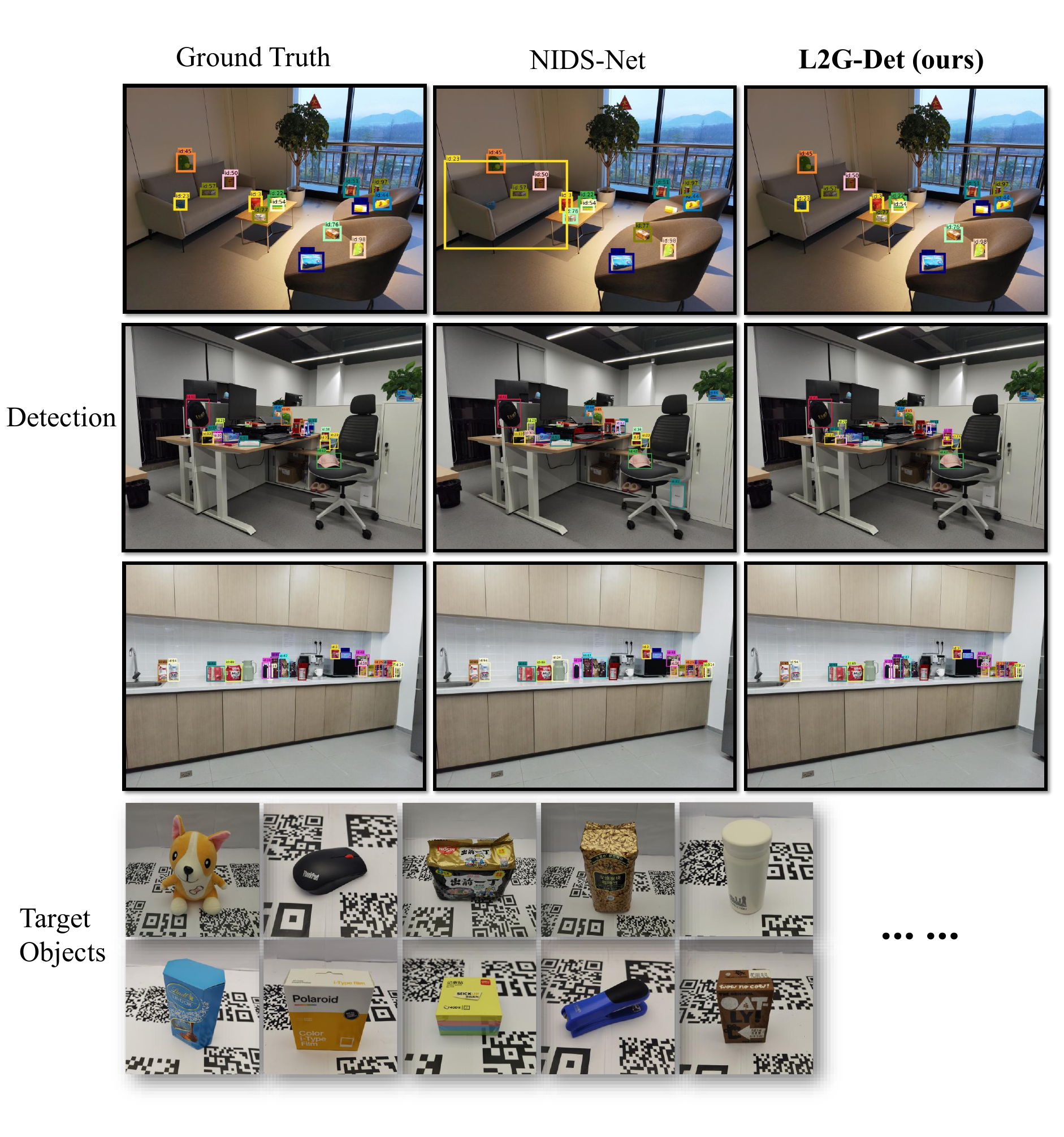}
  \caption{Qualitative results on HR-InsDet~\cite{shen2023instance} benchmark.
From left to right, we show the ground-truth annotations, results produced by NIDS-Net~\cite{lu2024adapting}, and results of our method \textbf{L2G-Det}.}
\vspace{-4mm}
  \label{fig:High_Res_results}
\end{figure*}

Qualitative comparisons on HR-InsDet~\cite{shen2023instance} are shown in Fig.~\ref{fig:High_Res_results}, where our method produces more complete and accurate detections in cluttered scenes.

\subsection{Template-based Synthetic Images}
\label{Appendix:Sync_images}

\begin{figure*}

  \centering
  \includegraphics[width=\linewidth]{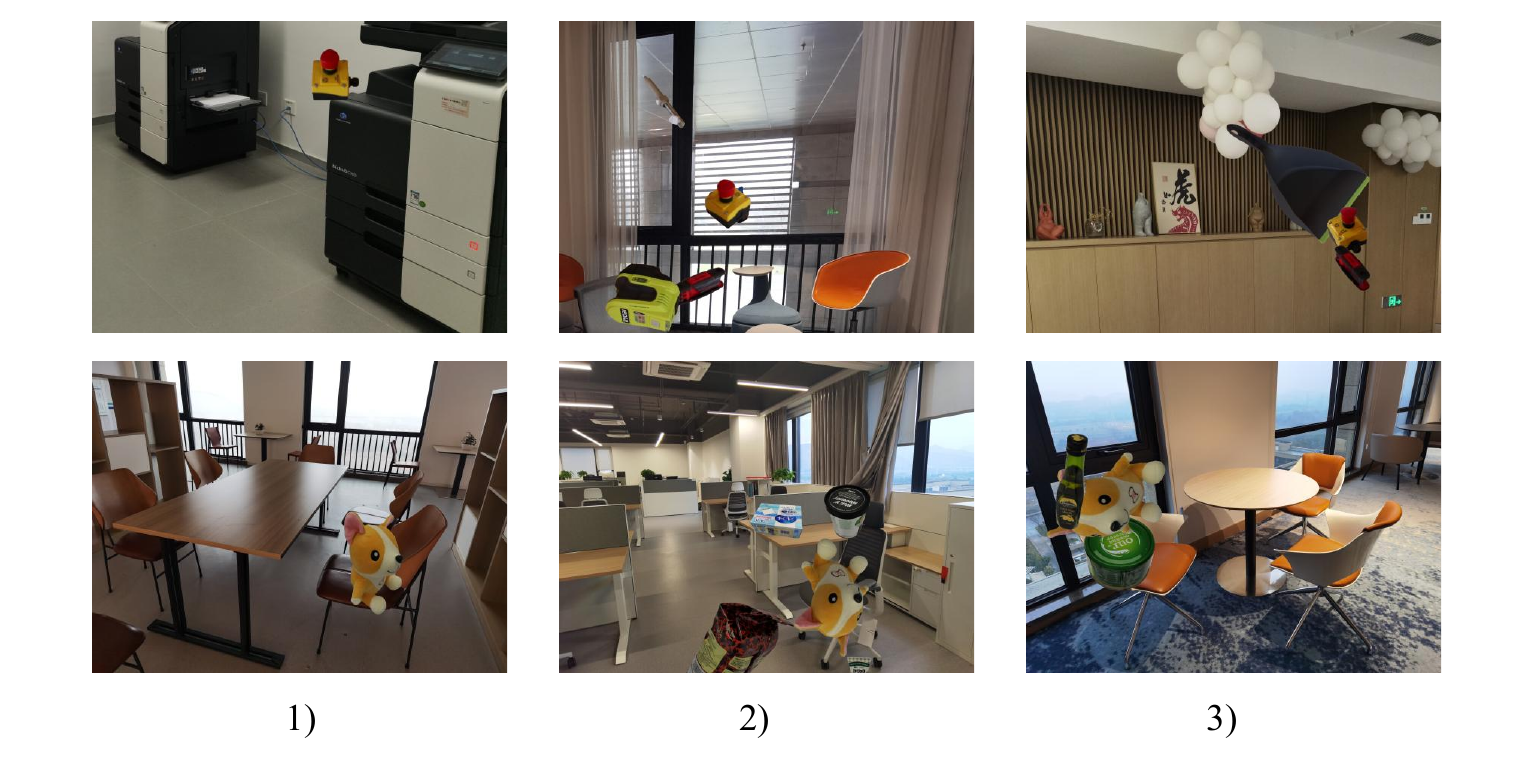}
  \caption{Examples of template-based synthetic training images.
Three synthesized scenes are illustrated: 1) Single-object composition, 2) Multi-object composition without overlap, and 3) Multi-object composition with overlap, where the target object may be partially occluded by surrounding objects.}
\vspace{-4mm}
  \label{fig:Syn_images}
\end{figure*}

We consider three types of synthesized scenes and show some examples in the Fig.~\ref{fig:Syn_images}:

\begin{enumerate}
    \item \textbf{Single-object composition:} only the target object is pasted onto the background image.
    \item \textbf{Multi-object composition without overlap:} the target object and several other objects are pasted onto the background, with additional objects placed around the target object without overlapping it.
    \item \textbf{Multi-object composition with overlap:} both the target object and other objects are pasted onto the background, where objects may partially overlap.
    When the target object is in the frontmost position, it remains unoccluded; otherwise, it may be partially occluded by surrounding objects, simulating challenging real-world conditions.
\end{enumerate}

\subsection{Real-world Robot Experiments}

\begin{figure*}[t]
    \centering
    \includegraphics[height=\textheight]{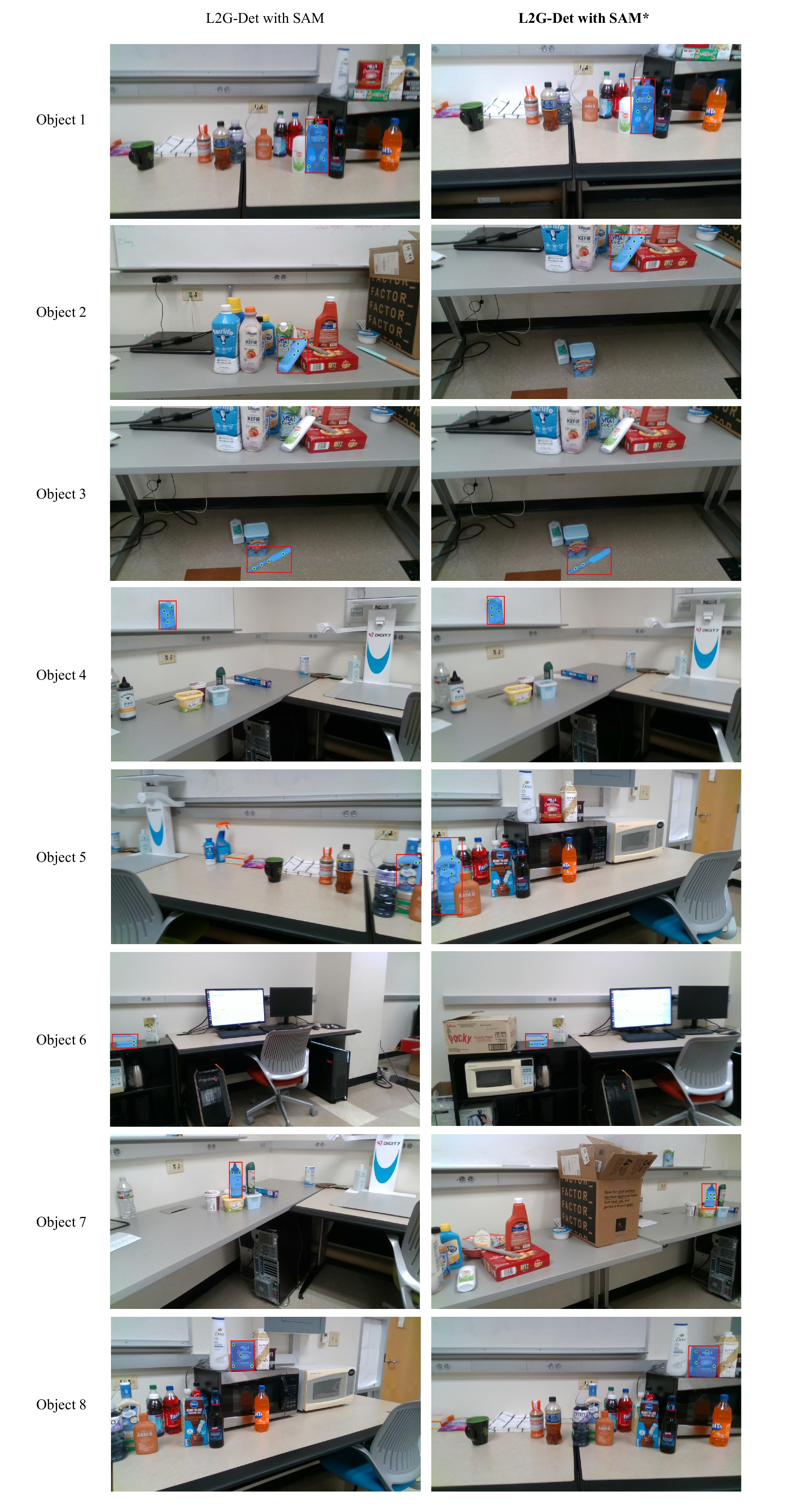}
    \caption{Qualitative real-world detection results on 8 target objects 
with and without instance-specific object tokens.}
    \label{fig:real_robot_results}
\end{figure*}

Fig.~\ref{fig:real_robot_results} shows the real-world detection results of 8 target objects under two settings:
(1) \textbf{L2G-Det with SAM}, and
(2) \textbf{L2G-Det with SAM$^{*}$}, which incorporates instance-specific object tokens.
For all 8 objects, the robot is able to successfully localize and detect the target instances in cluttered indoor environments.

Detailed navigation and detection processes using \textbf{Augmented SAM with instance-specific object tokens}
are provided in the supplementary video on our project website: \url{https://irvlutd.github.io/L2G/}.

\end{document}